\documentclass[dvipsnames]{article} %
\usepackage{colm2024_conference}

\usepackage{microtype}
\usepackage{hyperref}
\usepackage{url}
\usepackage{booktabs}
\usepackage{graphicx}
\usepackage{wrapfig}
\usepackage{bbm}
\usepackage{amsmath}
\usepackage{amssymb}
\usepackage{mathtools}
\usepackage{amsthm}
\usepackage{pifont}
\usepackage{footmisc}
\usepackage{enumitem}
\usepackage{natbib}
\usepackage{xcolor}
\usepackage{multirow}
\usepackage{setspace}
\usepackage{enumitem}
\usepackage{color, colortbl}
\usepackage{mathtools}
\usepackage{cleveref}
\theoremstyle{plain}

\usepackage{hyperref}
\usepackage{listings}
\usepackage{url}
\usepackage{algorithm}
\usepackage{algorithmic}
\usepackage{color,graphicx}
\usepackage{siunitx}
\usepackage{tabularx}
\usepackage{soul}
\usepackage{graphics} 
\usepackage{bookmark}
\usepackage[dvipsnames, svgnames, x11names]{xcolor}
\usepackage{hyperref} 
\usepackage{textcomp}
\usepackage{multirow}
\usepackage{nth}
\usepackage{indentfirst}
\usepackage{siunitx}
\usepackage{changepage}
\usepackage{bm}
\usepackage{setspace}
\usepackage{natbib}
\usepackage{multirow}
\usepackage{multicol}
\usepackage{colortbl}
\usepackage{booktabs}
\usepackage{amsmath}
\usepackage{xcolor}
\usepackage{amsthm}
\theoremstyle{remark}

\usepackage{graphicx}
\usepackage{subcaption}
\usepackage{wrapfig}
\usepackage{enumitem}

\usepackage[most]{tcolorbox}
\definecolor{darkmagenta}{rgb}{0.56, 0.0, 1.0}  

\hypersetup{colorlinks,linkcolor={blue},citecolor={darkmagenta},urlcolor={blue}}

\newtcolorbox{codeblock}{
  colback=gray!10,   
  colframe=gray!50,  
  boxrule=0.5mm,     
  arc=2mm,           
  left=5pt,          
  top=5pt,          
  bottom=5pt,        
}

\newtcbtheorem{proposition}{Proposition}{
  fonttitle=\bfseries,
  fontupper=\normalfont,
  boxrule=0.5pt,
  left=3mm,      
  right=4mm,
  top=2mm,
  bottom=2mm,
  clip upper=false,  
  breakable=false,
  before upper=\setlength{\parindent}{1em}
}{prop}

\newtcbtheorem{definition}{Definition}{
  enhanced,
  colback=cyan!5!white,           
  colframe=cyan!45!blue!60,       
  fonttitle=\bfseries,            
  fontupper=\normalfont,          
  arc=4mm,                        
  boxrule=1pt,
  drop shadow southwest,
  attach boxed title to top left={xshift=6mm, yshift=-2mm},
  boxed title style={
    colback=cyan,
    size=small,
    arc=2mm,                      
    boxrule=0.8pt,
    left=2mm, right=2mm
  },
  before skip=10pt,
  after skip=10pt,
  left=5mm,
  right=5mm,
  top=3mm,
  bottom=3mm
}{cor} 

\usepackage{amsmath,amsfonts,bm}

\def\eqref#1{equation~\ref{#1}}

\def\1{\bm{1}}

\DeclareMathAlphabet{\mathsfit}{\encodingdefault}{\sfdefault}{m}{sl}
\SetMathAlphabet{\mathsfit}{bold}{\encodingdefault}{\sfdefault}{bx}{n}

\renewcommand{\ghlink}{https://github.com/alibaba/ROLL}

\newcommand*\justify{%
  \fontdimen2\font=0.4em
  \fontdimen3\font=0.2em
  \fontdimen4\font=0.1em
  \fontdimen7\font=0.1em
  \hyphenchar\font=`\-
}

\renewcommand{\texttt}[1]{%
  \begingroup
  \ttfamily
  \begingroup\lccode`~=`/\lowercase{\endgroup\def~}{/\discretionary{}{}{}}%
  \begingroup\lccode`~=`[\lowercase{\endgroup\def~}{[\discretionary{}{}{}}%
  \begingroup\lccode`~=`.\lowercase{\endgroup\def~}{.\discretionary{}{}{}}%
  \catcode`/=\active\catcode`[=\active\catcode`.=\active
  \justify\scantokens{#1\noexpand}%
  \endgroup
}

\definecolor{blueviolet}{RGB}{138,43,226}

\newtcolorbox{abstractbox}{
    colback=blue!5!white,     
    frame empty,              
    boxrule=1pt,              
    arc=4mm,                  
    left=8pt,                 
    right=8pt,                
    top=8pt,                  
    bottom=8pt,                
    opacityback=0.9
}

\title{Complementary RL: Towards Efficient Experience-Driven Agent Learning}

\author{
{\bf 
\mbox{Dilxat Muhtar$^{1}$\thanks{Equal contribution}, Jiashun Liu$^{1,2*}$, 
Wei Gao$^{1,2}$, 
Weixun Wang$^{1}$, 
Shaopan Xiong$^{1}$, 
Ju Huang$^{1}$}, \\
\mbox{
Siran Yang$^{1}$,
Wenbo Su$^{1}$, 
Jiamang Wang$^{1}$, 
Ling Pan$^{2}$, 
Bo Zheng$^{1}$}}
\\
$^1$Alibaba Group \quad 
$^2$HKUST \quad
}

\author{
{\bf 
\mbox{Dilxat Muhtar$^{1,\dagger}$,
Jiashun Liu$^{1,2,\dagger}$, 
Wei Gao$^{1,2}$, 
Weixun Wang$^{1,\ast}$, 
Shaopan Xiong$^{1}$, 
Ju Huang$^{1,\ast}$}, \\
\mbox{
Siran Yang$^{1}$,
Wenbo Su$^{1}$, 
Jiamang Wang$^{1}$, 
Ling Pan$^{2}$, 
Bo Zheng$^{1}$}}
\\
$^1$Alibaba Group \quad 
$^2$HKUST \quad
}

\definecolor{cyan}{cmyk}{.3,0,0,0}

\begin{document}

\maketitle
\renewcommand{\thefootnote}{}
\footnotetext{$^\dagger$Equal contribution.}
\footnotetext{$^\ast$Corresponding authors: \url{weixun.wwx@taobao.com}; \url{huangju.hj@alibaba-inc.com}.}
\renewcommand{\thefootnote}{\arabic{footnote}}



\vspace{-4mm}
\begin{abstractbox}
\begin{center}
\vspace{-1mm}
\textbf{\Large Abstract}
\end{center}
Reinforcement Learning (RL) has emerged as a powerful paradigm for training LLM-based agents, yet remains limited by low sample efficiency, stemming not only from sparse outcome feedback but also from the agent's inability to leverage prior experience across episodes.
While augmenting agents with historical experience offers a promising remedy, existing approaches suffer from a critical weakness: the experience distilled from history is either stored statically or fail to co-evolve with the improving actor, causing a progressive misalignment 
between the experience and the actor's evolving capability that diminishes its utility over the course of training.
Inspired by complementary learning systems in neuroscience, 
we present \texttt{Complementary RL}
to achieve seamless co-evolution of an experience extractor and a policy actor within the RL optimization loop.
Specifically, the actor is optimized via sparse outcome-based rewards, while the experience extractor is optimized according to whether its distilled experiences demonstrably contribute to the actor's success, thereby evolving its experience management strategy in lockstep with the actor's growing capabilities.
Empirically, \texttt{Complementary RL} outperforms outcome-based agentic RL baselines that do not learn from experience, 
achieving 
$10\%$ performance improvement in single-task scenarios and exhibits robust scalability in multi-task settings. 
These results establish \texttt{Complementary RL} as a 
paradigm for efficient experience-driven agent learning.
\footnote{We release our training framework and training demo at \href{https://github.com/pUmpKin-Co/ComplementaryRL\#}{\texttt{here}}}

\end{abstractbox}
\begin{figure}[th]
  \centering
  \begin{subfigure}[t]{0.99\textwidth}
    \centering
    \vspace{0pt}
    \includegraphics[width=\linewidth]{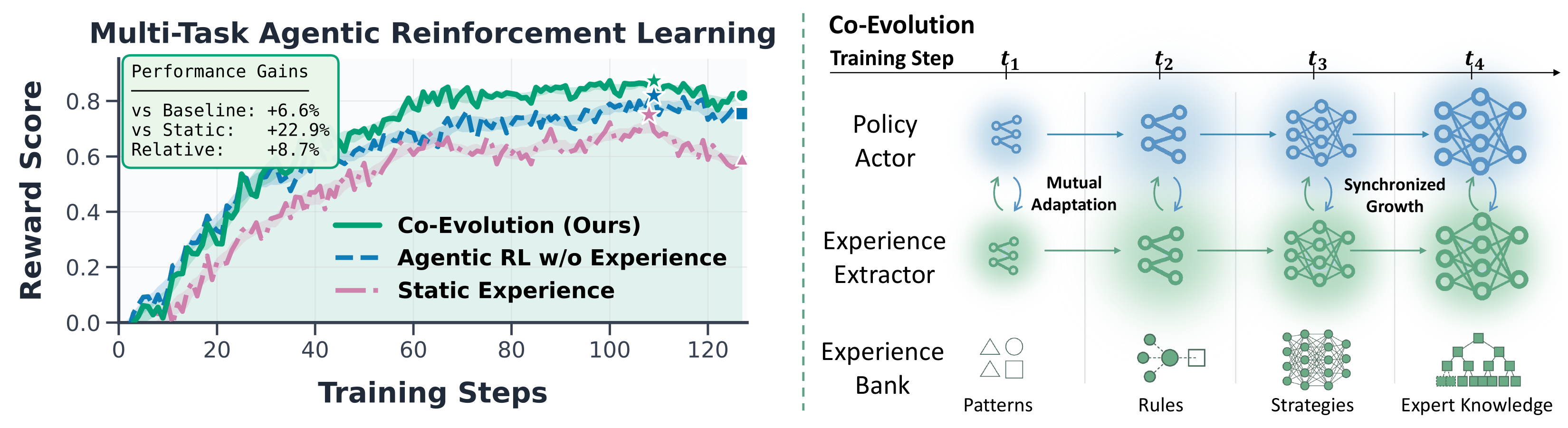}
    \vspace{-5mm}
  \end{subfigure}
  \vspace{-2mm}
  \caption{\texttt{Complementary RL} performance (left) and co-evolution paradigm (right).}
  \label{fig:leading_figure}
  \vspace{-5mm}
\end{figure}
\section{Introduction}
\vspace{-4mm}
Recent research has demonstrated the effectiveness of Reinforcement Learning (RL) in enhancing the agentic capabilities of Large Language Models (LLMs) \citep{jin2025search,dong2025agentic,xue2025simpletir}.
Despite this progress, outcome-based RL for LLMs-based agents remains limited by sample inefficiency.
Policy updates rely solely on sparse reward 
signals~\citep{shao2024deepseekmath, li2023remax, yu2025dapo}, which, while effective at optimizing task outcomes, provide no explicit signal for \textit{why} a trajectory 
succeeded or failed throughout the multi-turn interaction 
process~\citep{wang2026a}. 
Consequently, the rich procedural 
information embedded in collected rollouts, such as effective behaviors, recoverable failure patterns, and critical decision points, is largely unexploited. 
This underutilization of these procedural information renders the agent's learning process sample-inefficient \citep{zhang2026improving}.

To mitigate this inefficiency, 
a growing line of work explores how to leverage historical 
experience to increase the utilization of already-collected rollout data, therefore allowing the actor to learn fast~\citep{sutton2025}.
Here, we 
define \textit{experience} as structured textual knowledge distilled from raw trajectories, 
encompassing successful strategies, failure patterns, and generalizable decision rules.
A direct approach distills experience through self-generated reflections and incorporates 
it as in-context guidance during training~\citep{zhan2025exgrpo}.
However, when the base model is weak or tasks are complex, self-reflection becomes unreliable, frequently producing hallucinations that corrupt rather than enrich the learning signal~\citep{lin2025llm}.
To improve the reliability of experience used to guide the actor, some works focus on enhancing the quality of collected experience, 
either by maintaining auto-optimizing experience bank via specialized data 
structures~\citep{qian2025memorag, ouyang2025reasoningbank} or by employing a dedicated 
experience model to distill and dynamically refine structured experience from actor 
interactions~\citep{zhai2025agentevolver, zhang2025agent, xia2026skillrl, yan2025memory}. 
Others instead focus on designing multi-stage retrieval heuristics to surface the most valuable experience from the accumulated experience bank~\citep{zhou2025memento,zhang2026memrl}.



\begin{wrapfigure}{r}{0.6\textwidth}
    \begin{minipage}{0.6\textwidth}
        \centering  
        \vspace{-6mm}
        \centering
        \scalebox{1.00}
        {
            \begin{subfigure}[t]{\textwidth}
                \centering
                \includegraphics[width=\linewidth]{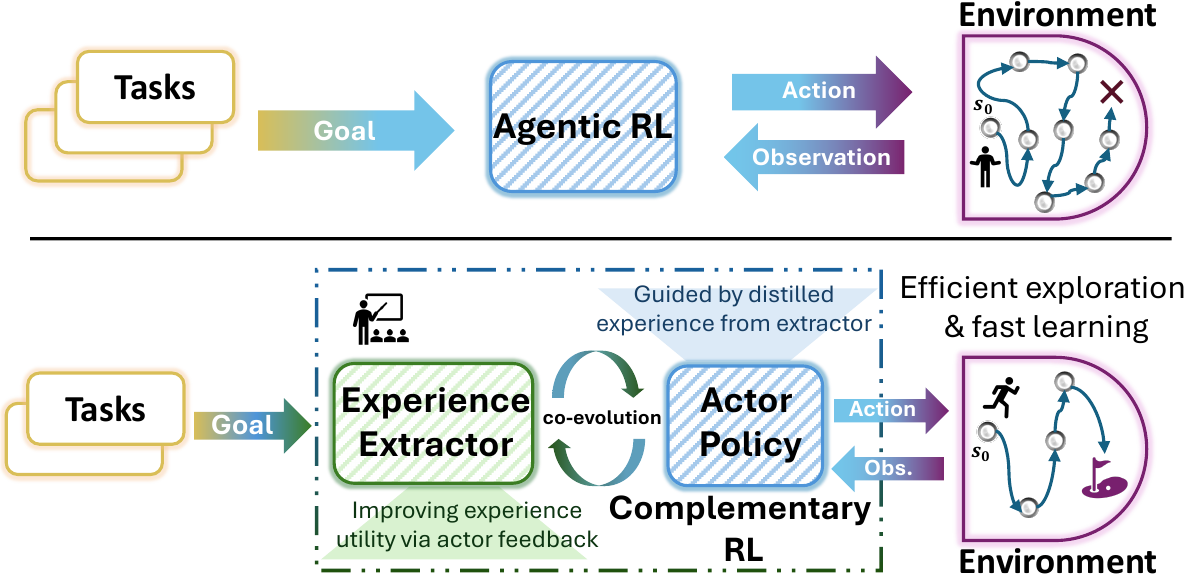}
            \end{subfigure}
        }
        \vspace{-6mm}
        \caption{Overview of \texttt{Complementary RL}.}
        \label{fig:algo_overview}
        \vspace{-4mm}
    \end{minipage}
\end{wrapfigure}
Despite the efforts to enable agents to learn from experience, prior 
works treat experience as a static resource, either maintaining fixed 
experience banks or 
employing non-adaptive experience extractors that progressively lag behind the actor's evolving capabilities, producing increasingly misaligned experience as training advances. 
Such stale experience limits learning efficiency as the actor grows stronger (Figure~\ref{fig:leading_figure} and Figure~\ref{sub_fig:formulaton}).
To improve the quality and relevance of experience throughout training, we argue that an RL algorithm for experience-driven agent training must satisfy three core design requirements:
\ding{182} \textbf{Actor-Extractor Co-Evolution:} the actor and 
experience extractor must mutually adapt throughout training, each 
continuously shaping the other toward greater capability;
\ding{183} \textbf{Experience Consolidation:} the experience 
bank must be automatically constructed and maintained from trajectories, 
distilling transferable experience while resolving conflicts and redundancies;
and \ding{184} \textbf{Training-Distillation Coordination:} actor training 
and experience distillation must be efficiently coordinated at scale without 
introducing blocking latency to actor training.
Motivated by these requirements, in this paper we aim to answer:

\textcolor{purple}{\textbf{Can we design a RL framework in which the policy 
actor and its experience extractor form a closed co-evolutionary loop, each 
continuously shaping the other toward better?}}


Interestingly, the human brain has long solved an analogous problem. 
Complementary Learning Systems (CLS) in neuroscience~\citep{OReilly2011-rv} enable the brain to rapidly acquire new knowledge while preserving long-term structured representations through two complementary systems:
the neocortex forms slow, structured long-term knowledge (analogous to the actor's policy), while the hippocampus manages fast, episode-specific memories (analogous to generated experiences), consolidating valuable episodes via cortical feedback and replaying them to strengthen decision-making.

Motivated by CLS, we propose \textbf{Complementary Reinforcement Learning} 
(\texttt{Complementary RL}), a RL algorithm built around two complementary 
models: an \textbf{actor} that interacts with the environment and 
optimizes guided by distilled experience, and an \textbf{experience 
extractor} 
responsible for distilling and maintaining a 
continuously evolving experience bank.
Both models are 
optimized via RL: the actor is trained using outcome-based rewards, while 
the extractor is optimized based on the utility of its distilled experience in facilitating the actor's success (Figure~\ref{fig:algo_overview}).
Through this mutual optimization, 
\texttt{Complementary RL} jointly meets 
the three requirements above: 
\ding{182} the actor and extractor form a closed 
co-evolutionary loop, where the extractor continuously refines experience to match the actor's growing capability and the actor benefits from increasingly 
relevant guidance;
\ding{183} the extractor distills experience from trajectories 
through structured addition, refining, and merging operations that 
automatically resolve conflicts and redundancies;
and 
\ding{184} We introduce a dedicated asynchronous training framework with a centralized experience manager that decouples actor interaction from experience distillation and dual-model optimization, ensuring training efficiency without introducing additional blocking latency.

\vspace{-1mm}
In summary, our main contributions are as follows:
\vspace{-1mm}
\definecolor{blueviolet}{RGB}{138,43,226}
\newtcolorbox{insightblock}{
  colback=blueviolet!5,   
  colframe=blueviolet!50!black!50!,    
  boxrule=0.5mm,       
  arc=2mm,            
  left=0pt,           
  right=8pt,           
  top=6pt,            
  bottom=6pt,}
\begin{insightblock}
\begin{enumerate}[leftmargin=1.5em]
    \item We propose \texttt{Complementary RL}, a paradigm that enables the co-evolution of a policy actor and an experience extractor during RL training, where the experience extractor continuously extracts and manages experience while the actor internalizes it to enable efficient policy improvement.~(\S\ref{sec:method})
    \item We develop an efficient training framework tailored for \texttt{Complementary RL}, featuring a fully asynchronous design 
    with a centralized \texttt{MemoryManager} that enables experience management 
    at scale.~(\S\ref{sec:pipeline}, \S\ref{appendix:impl_tricks})
    \item Through extensive empirical evaluation, we demonstrate the effectiveness of \texttt{Complementary RL}, and share key insights and lessons learned throughout the process.~(\S\ref{sec:experiments}, \S\ref{appendix:learned_lessons})
\end{enumerate}
\end{insightblock}
\section{Methodology}
\label{sec:method}
\subsection{Problem Formulation}
\label{subsec:preliminary}
We consider an LLM-based actor $\pi_{\theta}$ operating in an interactive environment 
$\mathcal{E}$, formalized as a Markov Decision Process (MDP) 
$\langle \mathcal{S}, \mathcal{A}, \mathcal{T}, \mathcal{R} \rangle$~\citep{NIPS2010_edfbe1af}, 
where $\mathcal{S}$, $\mathcal{A}$ are the state and action 
spaces, $\mathcal{T}: \mathcal{S} \times \mathcal{A} \rightarrow \mathcal{S}$ is the 
transition function, and $\mathcal{R}: \mathcal{S} \times \mathcal{A} \rightarrow \mathbb{R}$ 
is the reward function. At the beginning of each episode, the agent receives a task goal $g$. 
At each timestep $t$, it receives an observation $s_t \in \mathcal{S}$, produces an internal 
reasoning trace by reflecting on the current observation and interaction history, and then decides an action $a_t \sim \pi_{\theta}(\cdot \mid s_{\leq t}, g)$~\citep{yao2022react}. 
The environment then transitions to the next 
state $s_{t+1}$. 
An episode terminates upon task completion or upon reaching 
$T_{\max}$ steps, yielding a outcome reward $R \in \{0, 1\}$. The objective is to maximize 
the expected success rate across diverse tasks and environments:
\begin{equation}
    \mathcal{J}(\theta) = \mathbb{E}_{\mathcal{E},\, g,\, \tau \sim \pi_{\theta}} \left[ R(\tau) \right],
    \label{eq:objective}
\end{equation}
where $\tau = (s_0, a_0, s_1, a_1, \ldots, s_T)$ denotes the full interaction trajectory.

The formulation above treats each trajectory $\tau$ in isolation, optimizing $\pi_{\theta}$ 
solely from binary outcome rewards, leaving the rich behavioral information embedded in each trajectory unexploited.
A natural path toward greater learning efficiency is to distill structured experience $m$ 
from past trajectories, store it in an experience bank $\mathcal{M}$, and retrieve relevant 
entries to guide $\pi_{\theta}$ in subsequent episodes~\citep{sutton2025,ouyang2025reasoningbank, 
zhang2026memrl, zhai2025agentevolver}.
This augments the original objective (Equation~\ref{eq:objective}) to:
\begin{equation}
    \mathcal{J}(\theta) = \mathbb{E}_{\mathcal{E},\, g,\, m \sim \mathcal{M},\, \tau \sim \pi_{\theta}(\cdot \mid g, m)} \left[ R(\tau) \right].
    \label{eq:objective_extended}
\end{equation}
\subsection{From Static to Co-Evolutionary Experience}\label{subsec:bank}
\begin{figure}[t]
  \centering
      \begin{subfigure}[t]{0.24\textwidth}
        \centering
        \includegraphics[width=\linewidth]{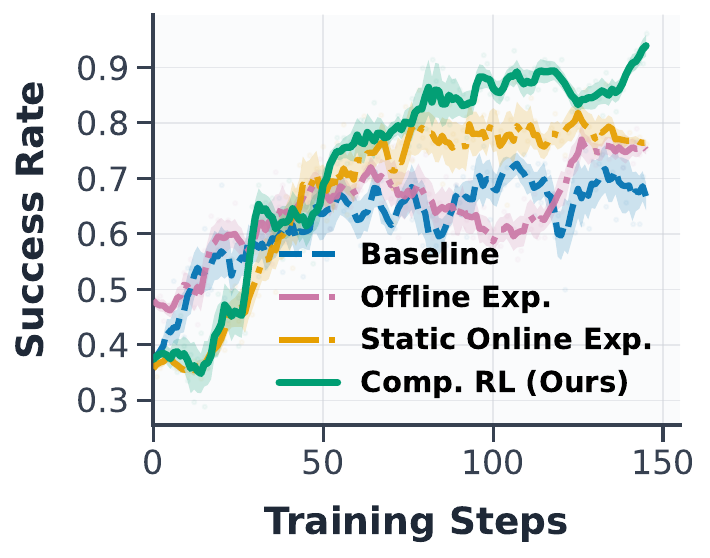}
        \vspace{-5mm}
        \caption{Exp. Comparison}
        \label{sub_fig:formulaton}
      \end{subfigure}
      \begin{subfigure}[t]{0.24\textwidth}
        \centering
        \includegraphics[width=\linewidth]{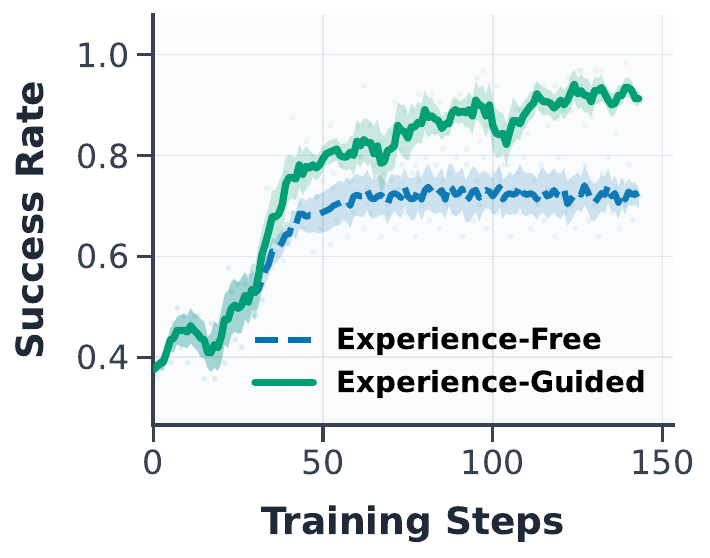}
        \vspace{-5mm}
        \caption{\texttt{w/o} Group Split}
        \label{sub_fig:full_group_rl}
      \end{subfigure}
      \begin{subfigure}[t]{0.24\textwidth}
        \centering
        \includegraphics[width=\linewidth]{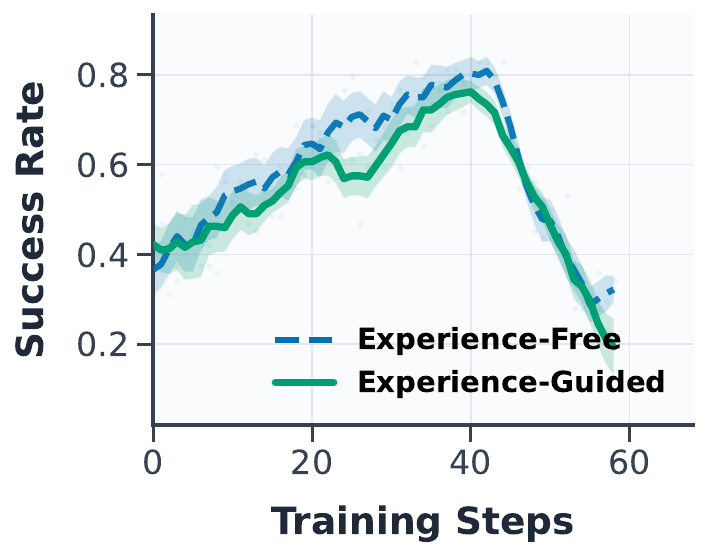}
        \vspace{-5mm}
        \caption{Cross-group Adv.}
        \label{sub_fig:split_group_rl}
      \end{subfigure}
      \begin{subfigure}[t]{0.24\textwidth}
        \centering
        \includegraphics[width=\linewidth]{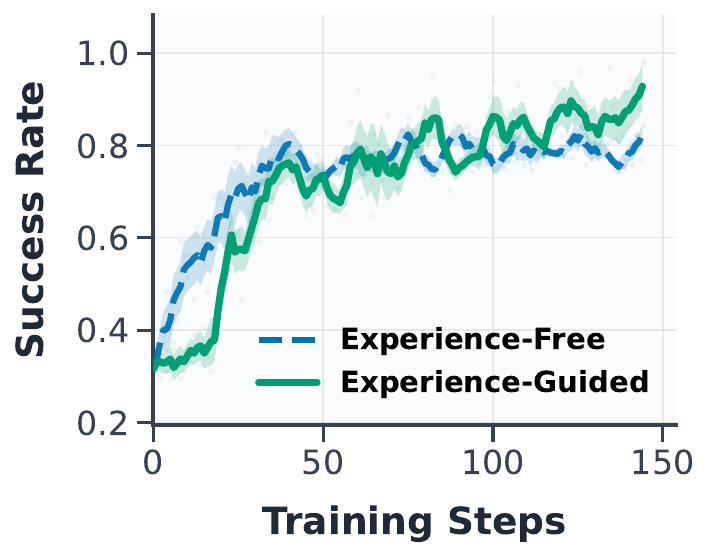}
        \vspace{-5mm}
        \caption{Subgroup Adv.}
        \label{sub_fig:split_group_diff_adv_rl}
      \end{subfigure}
      \vspace{-2mm}
        \caption{\textbf{(a)} Co-evolving the actor and experience extractor consistently outperforms static alternatives. \textbf{(b--d)} Ablation study on advantage estimation designs for the actor. The Exp.\ denotes experience.}
      \label{fig:method_motivation}
      \vspace{-3mm}
\end{figure}

Having formalized the learning-from-experience framework, we now turn to answering a practical question: how should the experience bank $\mathcal{M}$ be constructed and maintained to maximally benefit actor learning?
We analyze three design choices through a pilot study on the MiniHack Room~\citep{samvelyan2021minihack}
\footnote{\texttt{Room-Ultimate-5x5-v0}: \href{https://minihack.readthedocs.io/en/latest/envs/navigation/room.html}{\texttt{minihack-room}}}
: \textbf{(1) Baseline}: learning without experience; \textbf{(2) Offline Exp.}: $\mathcal{M}$ is pre-constructed from prior collected trajectories using an external  extractor~\citep{zhai2025agentevolver} and remains static during RL training; \textbf{(3) Static Online Exp.}: $\mathcal{M}$ is dynamically maintained by a fixed experience extractor $\pi_{\phi}$ during actor learning. 
Figure~\ref{sub_fig:formulaton} shows that while offline experience provides an initial performance boost, its benefit decays progressively over the course of training.
Similarly, static online experience yields only marginal gains over the baseline, suggesting that simply collecting online experience without co-evolving the extractor is insufficient.
We attribute this to a \textbf{distributional misalignment}: a static $\mathcal{M}$ cannot track the evolving state-action distribution of $\pi_{\theta}$, causing the guidance to become stale and counterproductive. 
This insight motivates us to the \textit{co-evolutionary} paradigm where $\pi_{\phi}$ and $\pi_{\theta}$ are jointly optimized. In this framework, improved policies generate higher-quality trajectories that refine $\mathcal{M}$, thereby providing more effective guidance for subsequent policy optimization. We formalize this mutually reinforcing mechanism as \texttt{Complementary RL}.

\subsection{Complementary Reinforcement Learning}
\label{sub_sec:comp_rl}
\paragraph{Algorithm Design for Experience Extractor}
In \texttt{Complementary RL}, the experience bank $\mathcal{M}$ is maintained by an experience extractor $\pi_{\phi}$, which is jointly optimized with the actor $\pi_{\theta}$. 
At the end of each episode, the extractor distills an experience entry $m \sim \pi_{\phi}(\cdot \mid g, \tau)$ conditioned on the task goal $g$ and the full interaction trace $\tau$. 
We track how $m$ influences subsequent actor behavior by assigning a binary reward $r(m) \in \{-1, +1\}$ based on the outcome of the trajectory it guided.
These experience-reward pairs are accumulated into a training batch $\mathcal{B}_{\phi} = \{(g_i, \tau_i, m_i, r(m_i))\}_{i=1}^O$, upon which $\pi_{\phi}$ is optimized via the CISPO objective \citep{chen2025minimax}:
\begin{equation}
    \mathcal{J}_{\text{CISPO}}(\phi) = \mathbb{E}\left[\frac{\displaystyle\sum_{i=1}^{O}\sum_{t=1}^{|m_i|)} \mathtt{sg}([\rho_{i,t}]^{1 + \varepsilon^{IS}_{high}}_{1 - \varepsilon^{IS}_{low}})\,\hat{A}_{i} \log\pi_{\phi}(m_{i,t} 
\mid g_i, \tau_i, m_{i,<t})}
    {\displaystyle\sum_{i=1}^{B}|m_i|}\right],\label{eq:cispo}
\end{equation}
where $\rho_{i,t} = \frac{{\pi_{\phi}(m_{i,t} \mid g_i, \tau_i, m_{i,<t})}}{{\pi_{\phi_{\mathrm{old}}}(m_{i,t} \mid g_i, \tau_i, m_{i,<t})}}$ is the token-level importance sampling (IS) ratio clipped to $[1 - \varepsilon^{IS}_{low},\, 1 + \varepsilon^{IS}_{high}]$.
$\mathtt{sg}(\cdot)$ denotes the stop-gradient operation, and $\hat{A}_{i} = r(m_i) - \bar{r}$ 
is the batch-level advantage, where $\bar{r}$ denotes the mean reward over batch $\mathcal{B}_{\phi}$, and $|m_i|$ denotes the number of tokens generated by $\pi_{\phi}$ for experience entry $m_i$.
We adopt CISPO instead of REINFORCE~\citep{sutton1999policy} to ensure stable co-evolution: the clipping mechanism constrains the IS ratio, preventing excessive policy updates that could cause the experience distribution to shift abruptly while ensuring that the gradients of all tokens are not wasted.

\paragraph{Algorithm Design for Actor} In practice, the actor $\pi_{\theta}$ is usually optimized via the GRPO~\citep{shao2024deepseekmath} objective, which maximizes the expected reward through group-relative advantage estimation over $K$ sampled trajectories $\{\tau_i\}_{k=1}^{K}$ per $(g, m)$:
\begin{equation}
    \mathcal{J}_{\text{GRPO}}(\theta) = \mathbb{E}\left[\frac{1}{K}\sum_{k=1}^{K} 
    \min\left( \rho \hat{A},\; 
    \text{clip}\left(\rho, 1-\varepsilon, 1+\varepsilon\right) \hat{A} \right) 
    \right],
    \label{eq:grpo}
\end{equation}
where $\rho = \frac{\pi_{\theta}(\tau \mid g, m)}{{\pi_{\theta_{\mathrm{old}}}(\tau \mid g, m)}}$ 
is the sequence level IS ratio, $\hat{A} = ({r(\tau) - \bar{r}})/{\sigma}$ is the group-normalized advantage, and $\varepsilon$ is the clipping threshold.

However, we observe that when \textit{all} interactions are conditioned on retrieved experience, the actor converges prematurely and lags behind the experience-guided setting (Figure~\ref{sub_fig:full_group_rl}), suggesting that the actor fails to internalize experience into its own capabilities and instead develops an over-reliance on external guidance. 
Inspired by~\citet{zhai2025agentevolver}, we therefore partition the $K$ rollouts evenly into two subgroups: \textbf{experience-guided} and \textbf{experience-free}. 
However, a critical issue arises when computing advantages across the two subgroups: the reward scales and variances differ between subgroups, causing advantage estimates to become biased and training to collapse (Figure~\ref{sub_fig:split_group_rl}). 
To preserve signal integrity, we propose computing advantages \textit{within} each subgroup, ensuring that relative performance is evaluated under consistent conditioning:
\begin{equation}
    \mathcal{J}_{\text{GRPO}}^{\text{split}}(\theta) = \mathbb{E}\!\left[
    \frac{1}{2}\sum_{c \in \{m, \varnothing\}}
    \frac{1}{K_c}\sum_{k=1}^{K_c}
    \mathcal{L}_{\text{clip}}\!\left(\rho_c, \hat{A}_c\right)
    \right],
    \label{eq:grpo_split}
\end{equation}
where $c \in \{m, \varnothing\}$ indexes the subgroup with experience-guided and experience-free interactions, and $\hat{A}_c = ({r(\tau_c}) - \bar{r}_c)/{\sigma_{c}}$ is normalized \textit{within} subgroup $c$ using its own mean $\bar{r}_c$ and standard deviation $\sigma_{c}$. 
In practice, the two subgroups are of equal size $K_c = K/2$, which ensures balanced gradient contributions from both two subgroups and prevents either condition from dominating the training signal. $\mathcal{L}_{\text{clip}}(\rho, A) = \min\left(\rho A, \text{clip}(\rho, 1-\varepsilon, 1+\varepsilon) A\right)$ is the clipped surrogate loss.
This condition-wise advantage estimation preserves the distinct learning signals of each condition and stabilizes training, yielding consistent improvement across both subgroups (Figure~\ref{sub_fig:split_group_diff_adv_rl}).
\section{Training Framework}
\label{sec:pipeline}
\begin{figure}[t]
  \centering
      \begin{subfigure}[t]{0.95\textwidth}
        \centering
        \includegraphics[width=\linewidth]{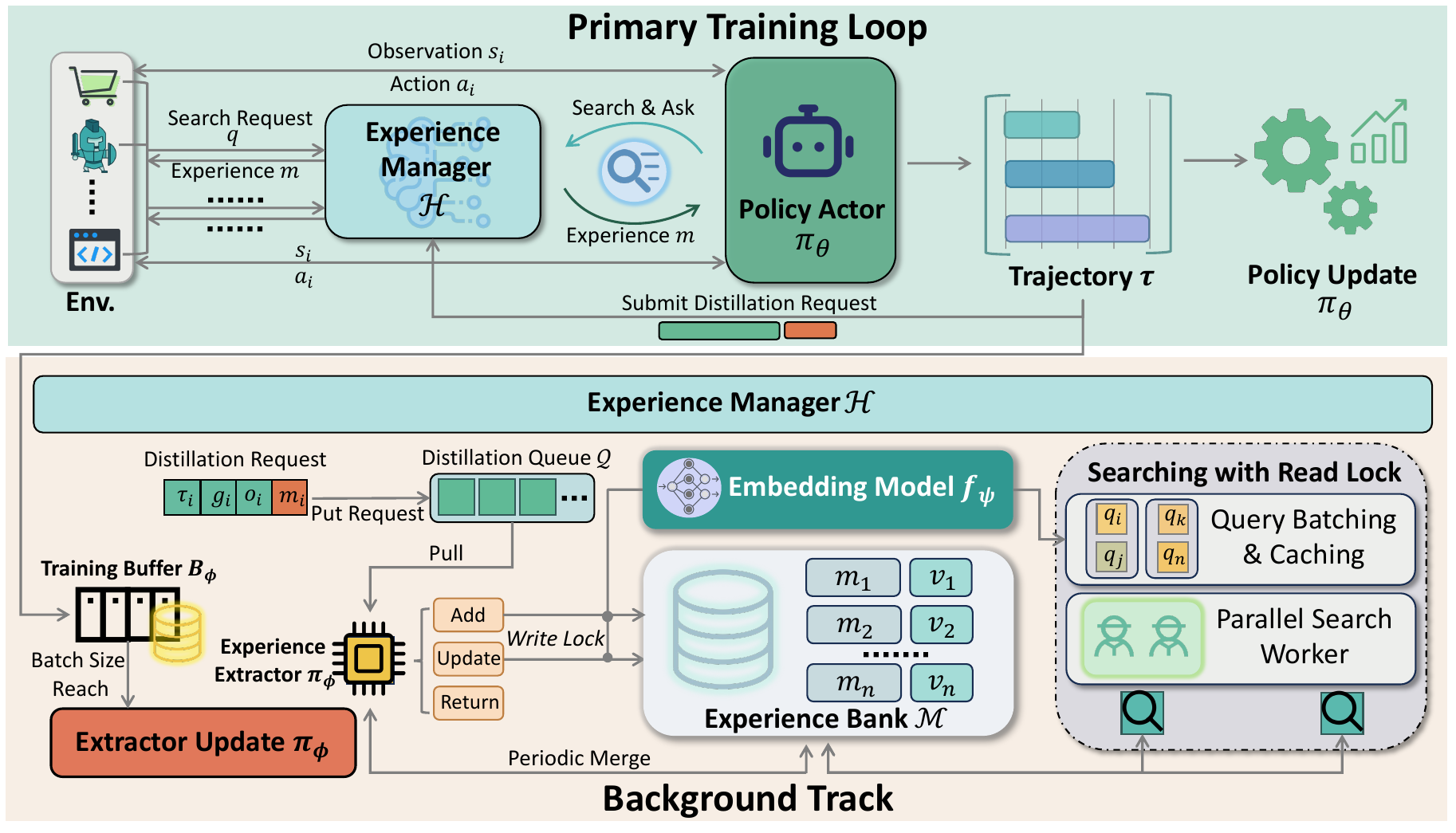}
      \end{subfigure}
      \vspace{-2mm}
       \caption{Overview of the \texttt{Complementary RL} training infrastructure, 
        where the actor and experience extractor are trained asynchronously and 
        coordinated through a centralized experience manager.}
      \label{fig:pipeline_overview}
      \vspace{-6mm}
\end{figure}
\subsection{Overview}
\texttt{Complementary RL} jointly optimizes the policy actor $\pi_{\theta}$ and the experience extractor $\pi_{\phi}$, where the two models are mutually dependent: $\pi_{\theta}$ requires retrieved experience before each interaction, while $\pi_{\phi}$ depends on completed actor trajectories for distillation and receives training signals reflecting whether the experience it produced was beneficial.
A naïve implementation would serialize these dependencies, where after each batch of rollouts, actor training would block while waiting for experience distillation and $\pi_{\phi}$ optimization to complete, introducing synchronization barriers that cause significant resource idleness and degrade overall training throughput.

To eliminate this bottleneck, \texttt{Complementary RL} deliberately decouples rollout collection from experience distillation via a fully asynchronous design comprising a \textbf{primary training loop} and a \textbf{background track}, as illustrated in Figure~\ref{fig:pipeline_overview}.
In the \textbf{primary training loop}, the actor $\pi_{\theta}$ continuously interacts with the environment to collect rollouts and is optimized via outcome-based rewards.
Concurrently, in the \textbf{background track}, the experience extractor $\pi_{\phi}$ processes completed trajectories, distills experience, and issues structured operations to maintain the experience bank $\mathcal{M}$.

Although the two tracks run asynchronously, they remain tightly coupled: 
at the beginning of each episode, relevant experience is retrieved from $\mathcal{M}$ to condition $\pi_{\theta}$, and upon episode completion, regardless of success or failure, the full trajectory is forwarded to $\pi_{\phi}$ for distillation.
Coordinating these interactions at scale, where hundreds of environments execute in parallel while sharing a single globally consistent $\mathcal{M}$, requires careful concurrency management.
To this end, we introduce a centralized \texttt{ExperienceManager} $\mathcal{H}$, which serves two coordinating roles:
\textbf{(1) Experience Consolidation:} $\mathcal{H}$ maintains an internal queue $\mathcal{Q}$ to receive and schedule distillation requests, and manages all writes to $\mathcal{M}$ under a writer lock to prevent state conflicts (\S\ref{subsub_sec:consolidation});
\textbf{(2) Experience Retrieval:} $\mathcal{H}$ aggregates concurrent retrieval queries into micro-batches to maximize throughput, and distributes semantic search across parallel workers under a reader lock to enable concurrent reads (\S\ref{subsub_sec:retrieval}).
Through $\mathcal{H}$, \texttt{Complementary RL} achieves efficient experience management at scale, keeping the additional latency introduced to the actor training loop minimal.
In the following, we detail our infrastructure design for experience consolidation, retrieval, 
and co-evolution of $\pi_{\theta}$ and $\pi_{\phi}$, with additional stabilization tricks 
deferred to Appendix~\ref{appendix:impl_tricks}.

\subsection{Experience Consolidation and Retrieval}
\label{sub_sec:consolidation_retrieval}

\subsubsection{Experience Consolidation}
\label{subsub_sec:consolidation}

\paragraph{Producer-Consumer Distillation}
Upon completion of each episode, regardless of outcome, the full interaction trace 
$\tau$, together with the initial task goal $g$, 
the final outcome $o \in \{\text{success}, \text{failure}\}$, and the experience entry 
$m \in \mathcal{M}$ retrieved to guide the episode, are submitted to $\mathcal{H}$ as 
a distillation request.
$\mathcal{H}$ maintains an internal queue $\mathcal{Q}$ to receive distillation requests from all parallel environments. 
A background process continuously dequeues pending requests 
and forwards them to the experience extractor $\pi_{\phi}$ for distillation. 

For each distillation request $\mathcal{R} = (\tau, g, o, m)$, $\pi_{\phi}$ reasons over the full interaction trace, the episode outcome, and how the retrieved experience $m$ influenced the actor's behavior, before issuing the following structured operations: \texttt{Add} a 
newly synthesized experience entry into $\mathcal{M}$, \texttt{Update} the previously 
retrieved entry $m$, or \texttt{Return} without action when the episode yields no extractable insight.
Upon receiving the issued operations from $\pi_{\phi}$, $\mathcal{H}$ applies them to 
$\mathcal{M}$ under a writer lock, which temporarily suspends concurrent reads to prevent state conflicts. 
For each newly added experience entry $m$, it is first passed through an embedding model $f_{\psi}$ to obtain its dense vector $\mathbf{v}_m = f_{\psi}(m)$. 
The entry $m$, its embedding $\mathbf{v}_m$, and the generation prompt-response pair produced 
by $\pi_{\phi}$ are then jointly persisted to $\mathcal{M}$, enabling both semantic retrieval and future evolving of $\pi_{\phi}$.

\paragraph{Periodic Merge}
The above consolidation process treats each episode independently. 
However, in group-based RL, multiple instances of the same task typically run in parallel, which 
can lead to redundant or conflicting experience entries being added to $\mathcal{M}$. 
Such redundancy degrades the quality of semantic retrieval and consequently 
impairs the actor's learning (Figure~\ref{sub_fig:ablation_wo_merge}).
To mitigate this, we periodically trigger a \texttt{Merge} operation every several actor updates. 
Experiences in $\mathcal{M}$ are processed in chunks, each passed to $\pi_{\phi}$ with a structured prompt that instructs the model to analyze the semantic relationships among entries and decide which to retain, which to merge, and which to discard. 
The merged output is then carried forward and concatenated with the next chunk, forming a chunk-wise sliding process over the full $\mathcal{M}$. 
This design bounds the 
context length presented to $\pi_{\phi}$ while ensuring all entries are considered, yielding a compact experience bank that benefits 
actor learning.

\subsubsection{Experience Retrieval}
\label{subsub_sec:retrieval}

\paragraph{Query Batching and Parallel Search}
At the beginning of each episode, the environment submits a \texttt{Search} request to 
$\mathcal{H}$ using the task description as a query $q$. Rather than processing queries 
individually, $\mathcal{H}$ accumulates incoming queries into a waiting buffer until either 
a predefined batch size $B$ or a maximum waiting time $t_{\max}$ is reached. Each query 
is then checked against an embedding cache $\mathcal{C}$ before invoking $f_{\psi}$, which 
is particularly effective in group-based RL training where many parallel environments share 
identical task descriptions. Cache misses are forwarded to $f_{\psi}$ for batched embedding 
computation, yielding $\mathbf{v}_q = f_{\psi}(q)$. 
The resulting embeddings are distributed 
via round-robin to one of $W$ parallel search workers, each performing semantic similarity search over $\mathcal{M}$ under a reader lock, allowing concurrent reads while blocking writes. 
Finally, the most relevant 
experience entry $m$ is then returned to the requesting environment.
Through batching, caching, and parallel search, this design maximizes retrieval throughput while minimizing latency introduced to the actor's environment interaction.

\begin{wrapfigure}{r}{0.6\textwidth}
    \begin{minipage}{0.6\textwidth}
        \centering  
        \vspace{-5mm}
        \centering
        \scalebox{1.00}
        {
            \begin{subfigure}[t]{0.5\textwidth}
                \centering
                \includegraphics[width=\linewidth]{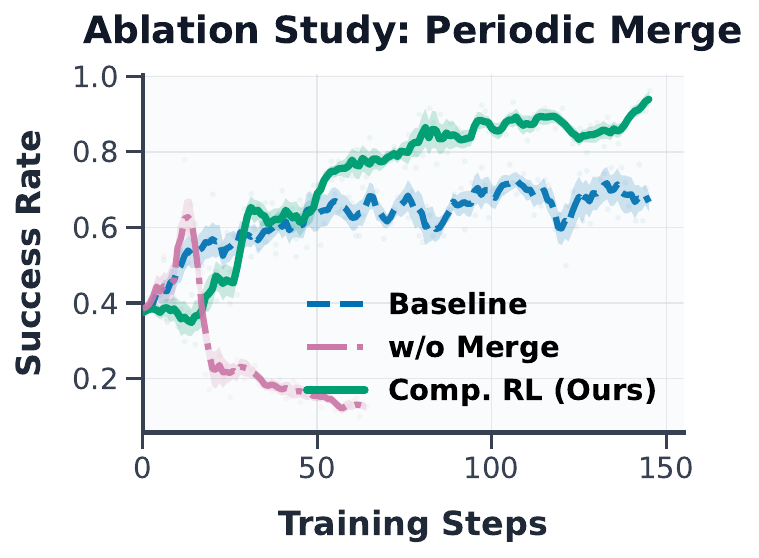}
                \vspace{-6mm}
                \caption{\textit{w/o} \texttt{Merge}}
                \label{sub_fig:ablation_wo_merge}
            \end{subfigure}
            \begin{subfigure}[t]{0.5\textwidth}
                \centering
                \includegraphics[width=\linewidth]{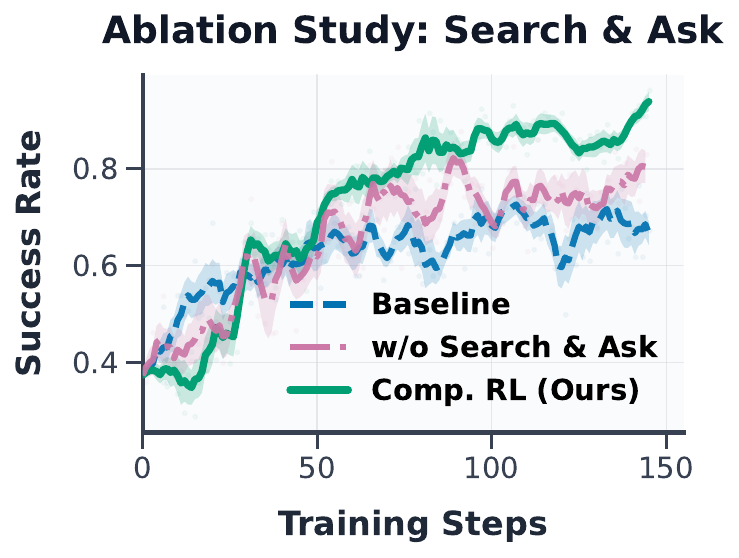}
                \vspace{-6mm}
                \caption{\textit{w/o} 
                \label{sub_fig:ablation_wo_tool}
                \texttt{search\_and\_ask}}
            \end{subfigure}
        }
        \vspace{-2mm}
        \caption{Ablation on \texttt{Merge} and \texttt{search\_and\_ask} in MiniHack Room.}
        \label{fig:ablation_merge_and_tool}
        \vspace{-3mm}
    \end{minipage}
\end{wrapfigure}
\paragraph{Search-and-Ask}
Using the task description alone as query $q$ tends to retrieve the same experience entry $m$ repeatedly, since parallel environments in group-based RL training often share identical 
task descriptions or differ only in environment-specific details such as map layouts (e.g., MiniHack~\citep{samvelyan2021minihack}). 
This reduces the utilization of $\mathcal{M}$ 
and limits the diversity of training signal available for optimizing $\pi_{\phi}$.
To address this, we introduce the \texttt{search\_and\_ask} tool, which allows $\pi_{\theta}$ 
to actively query $\mathcal{M}$ at any decision step during environment interaction. 
When the actor invokes this tool, it constructs a context-aware query $q'$ by summarizing its current state and the difficulties it faces, and submits $q'$ to $\mathcal{H}$ for retrieval.
If a relevant entry $m$ is found, the pair $(q', m)$ is 
forwarded to $\pi_{\phi}$, which refines $m$ according to the actor's specific situation before returning the result.
This mechanism increases $\mathcal{M}$ utilization, enriches the training signal for $\pi_{\phi}$, and enables the actor to obtain more targeted guidance aligned with its current situation at critical decision points, further improving learning efficiency (Figure~\ref{sub_fig:ablation_wo_tool}).

\subsection{Co-Evolution Training}
\label{sec:coevolution}

The actor $\pi_{\theta}$ is evolved following the objective described in Equation~\ref{eq:grpo_split}. 
For the evolution of $\pi_{\phi}$, after each rollout collection step that yields a batch of 
trajectories $\mathcal{T} = \{\tau_i\}_{i=1}^{N}$ for training $\pi_{\theta}$, we extract 
the experience entry $m$ that guided each trajectory $\tau_i$ and assign it a binary reward 
$r(m) \in \{-1, 1\}$ based on whether the corresponding episode succeeded. The prompt-response 
pair generated by $\pi_{\phi}$ to produce $m$ is then stored in a training buffer 
$\mathcal{B}_{\phi}$.
Since multiple trajectories in $\mathcal{T}$ may share the same retrieved entry $m$, we treat 
each unique $m$ as a single training sample and accumulate its rewards across all associated 
trajectories, assigning the average reward $\bar{r}(m) = \frac{1}{|\mathcal{T}_m|} 
\sum_{\tau \in \mathcal{T}_m} r(m, \tau)$, where $\mathcal{T}_m \subseteq \mathcal{T}$ 
denotes the subset of trajectories guided by $m$. As a result, the number of unique training 
samples for $\pi_{\phi}$ may be smaller than defined batch size for $\pi_{\phi}$, and a single rollout collection step may 
not suffice to fill $\mathcal{B}_{\phi}$. We therefore accumulate samples across multiple 
rollout collection steps, and only trigger the optimization of $\pi_{\phi}$ once 
$|\mathcal{B}_{\phi}|$ reaches the required training batch size, as described in 
Equation~\ref{eq:cispo}. 
Crucially, $\pi_{\phi}$ and $\pi_{\theta}$ are optimized on fully independent schedules, ensuring 
neither blocks nor interferes with the other throughout co-evolution training.
\section{Experiments}
\label{sec:experiments}
\subsection{Experimental Settings}
We evaluate \texttt{Complementary RL} on four open-ended environments: 
MiniHack~\citep{samvelyan2021minihack}, 
WebShop~\citep{yao2023webshopscalablerealworldweb}, 
ALFWorld~\citep{shridhar2021alfworldaligningtextembodied}, and 
SWE-Bench~\citep{jimenez2024swebenchlanguagemodelsresolve}. During training, 
we track success rate on MiniHack and WebShop, and reward on held-out 
evaluation sets for ALFWorld and SWE-Bench. 
For a fair comparison of final performance, all methods are  evaluated on fixed evaluation 
tasks for all environments. 
Detailed environment descriptions are provided in the Appendix~\ref{appendix:env_description}.

Without other specification, we use \texttt{Qwen2.5-7B-Instruct}~\citep{qwen2025qwen25technicalreport} as actor $\phi_{\theta}$ and use \texttt{Qwen3-4B-Thinking-2507}~\citep{yang2025qwen3technicalreport} as the experience extractor $\phi_{\phi}$.
For all of the comparison methods, we use the same hyperparameters for fail comparison, which we defer to Appendix~\ref{appendix:training_configuration} for detail introduction.

\begin{figure}[t]
  \centering
      \begin{subfigure}[t]{0.24\textwidth}
        \centering
        \includegraphics[width=\linewidth]{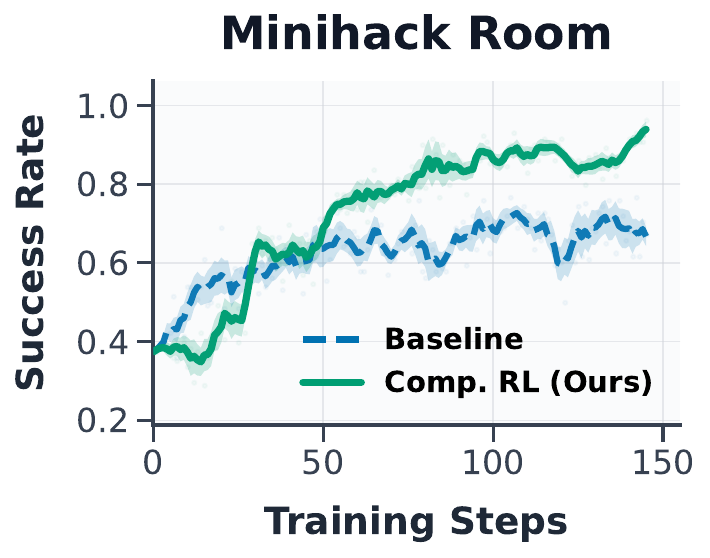}
        \vspace{-5mm}
      \end{subfigure}
      \begin{subfigure}[t]{0.24\textwidth}
        \centering
        \includegraphics[width=\linewidth]{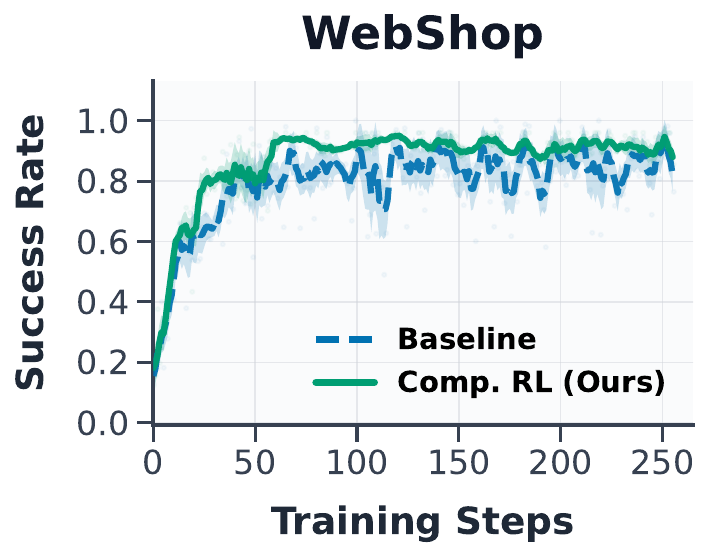}
        \vspace{-5mm}
      \end{subfigure}
      \begin{subfigure}[t]{0.24\textwidth}
        \centering
        \includegraphics[width=\linewidth]{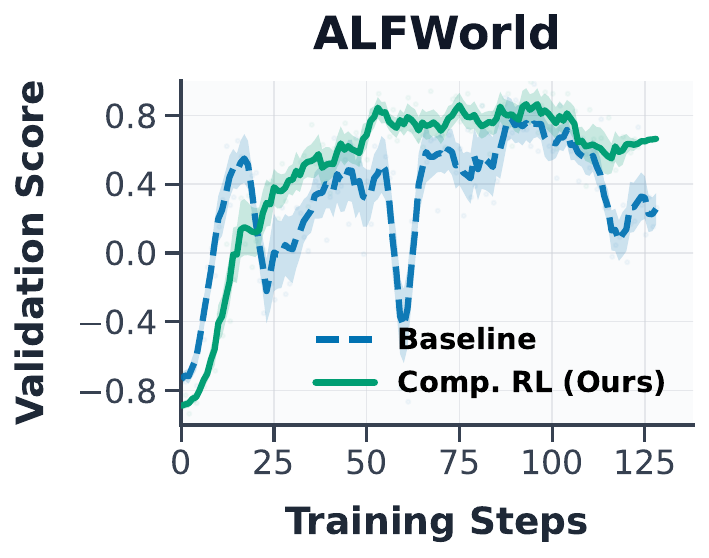}
        \vspace{-5mm}
      \end{subfigure}
      \begin{subfigure}[t]{0.24\textwidth}
        \centering
        \includegraphics[width=\linewidth]{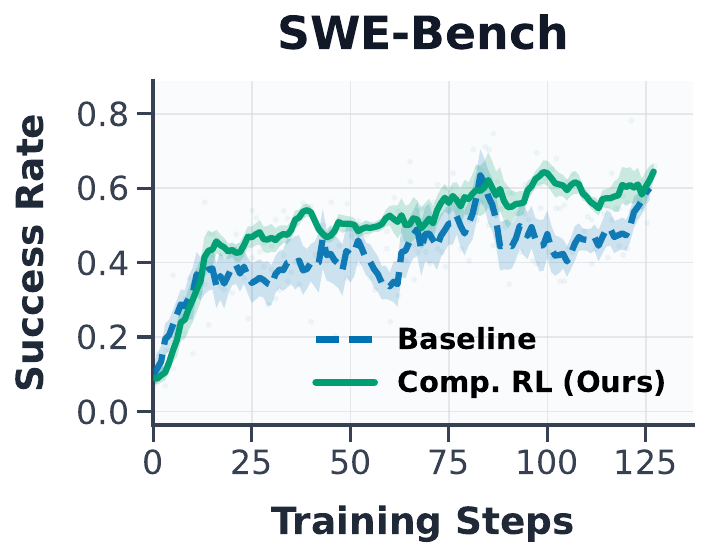}
        \vspace{-5mm}
      \end{subfigure}
      \vspace{-3mm}
        \caption{
        Single-task evaluation scores across four different environments.
        }
        \vspace{-3mm}
        \label{fig:main_result_singletask}
\end{figure}

\begin{figure}[t]
  \centering
      \begin{subfigure}[t]{0.24\textwidth}
        \centering
        \includegraphics[width=\linewidth]{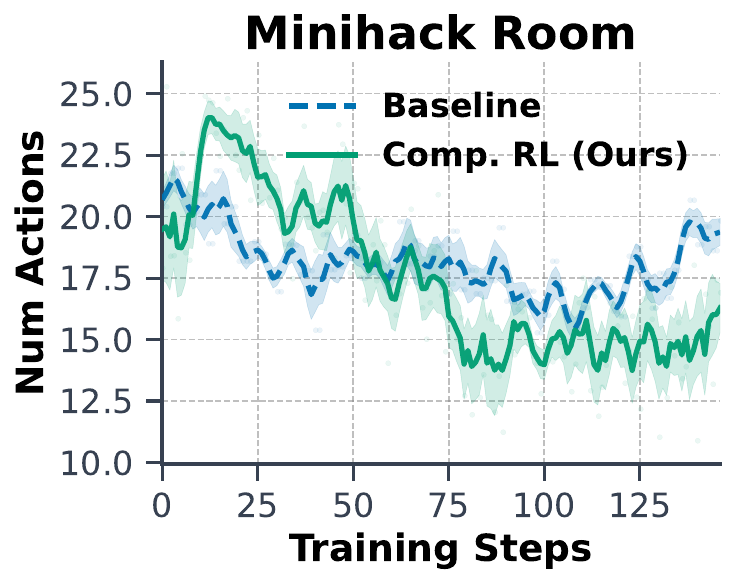}
        \vspace{-5mm}
      \end{subfigure}
      \begin{subfigure}[t]{0.24\textwidth}
        \centering
        \includegraphics[width=\linewidth]{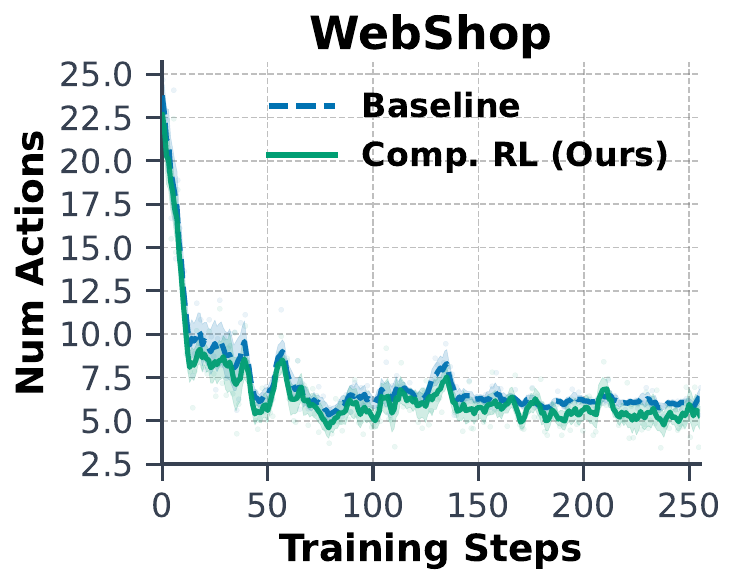}
        \vspace{-5mm}
      \end{subfigure}
      \begin{subfigure}[t]{0.24\textwidth}
        \centering
        \includegraphics[width=\linewidth]{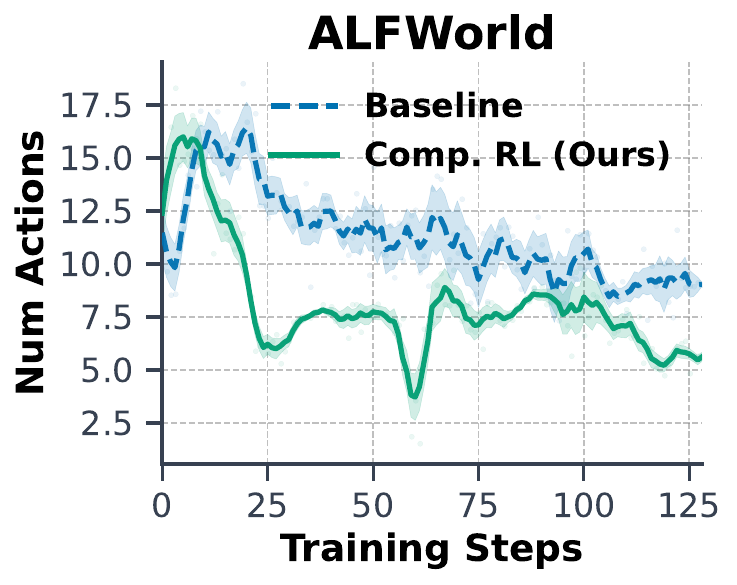}
        \vspace{-5mm}
      \end{subfigure}
      \begin{subfigure}[t]{0.24\textwidth}
        \centering
        \includegraphics[width=\linewidth]{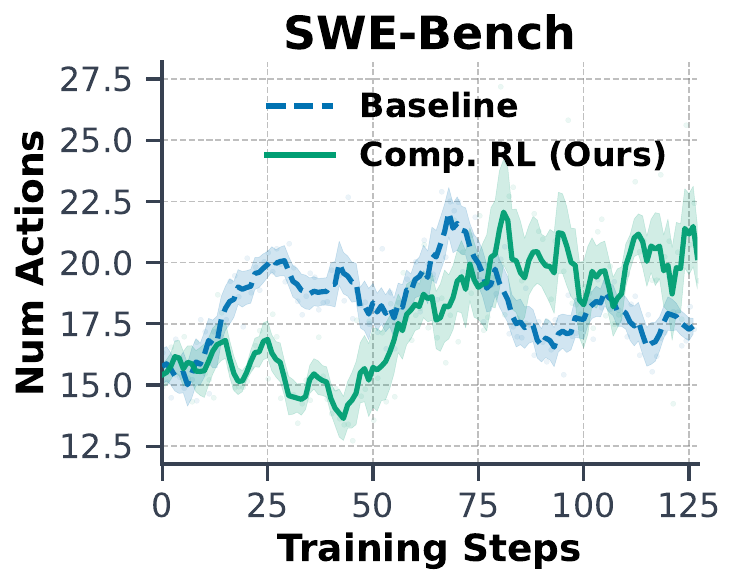}
        \vspace{-5mm}
      \end{subfigure}
      \vspace{-3mm}
        \caption{
        Average number of actions per task.
        }
        \vspace{-6mm}
        \label{fig:num_actions_singletask}
\end{figure}

\subsection{Main Result}
\paragraph{Single-Task Training}
We first evaluate \texttt{Complementary RL} separately on each of the four tasks and compare it against baselines that do not leverage experience. We use \texttt{Qwen3-4B-Instruct-2507} as the actor $\pi_{\theta}$ for SWE-Bench in this experiment, while all other tasks follow the default settings described earlier.

As shown in Figure~\ref{fig:main_result_singletask}, \texttt{Complementary RL} consistently outperforms the baseline across all four tasks.
In tasks requiring strategic exploration and environmental understanding, such as MiniHack Room and ALFWorld, \texttt{Complementary RL} achieves a $1.3\times$ performance margin with notably better training stability.
Moreover, on the challenging software engineering benchmark SWE-Bench, \texttt{Complementary RL} demonstrates faster improvement and achieves a $+3.0\%$ gain over the baseline.
Furthermore, Figure~\ref{fig:num_actions_singletask} reveals that \texttt{Complementary RL} not only achieves higher success rates but also completes tasks more efficiently, requiring $1.5\times$ fewer actions on MiniHack Room and $2\times$ fewer actions on ALFWorld, demonstrating that distilled experience guides the actor toward more effective decision-making.
Although \texttt{Complementary RL} exhibits an increasing number of actions on SWE-Bench, we find that this is because the agent takes more actions to fully complete tasks, thereby achieving a higher success rate, rather than submitting prematurely before a task is finished.

\begin{figure}[t]
  \centering
      \begin{subfigure}[t]{0.24\textwidth}
        \centering
        \includegraphics[width=\linewidth]{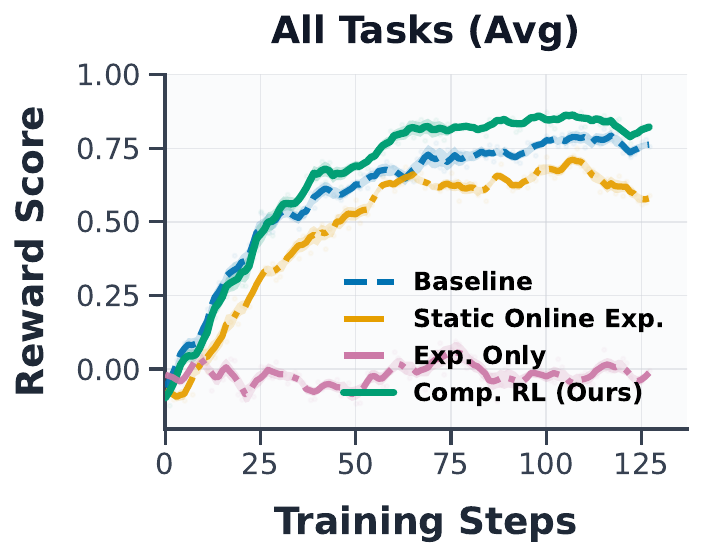}
        \vspace{-5mm}
      \end{subfigure}
      \begin{subfigure}[t]{0.24\textwidth}
        \centering
        \includegraphics[width=\linewidth]{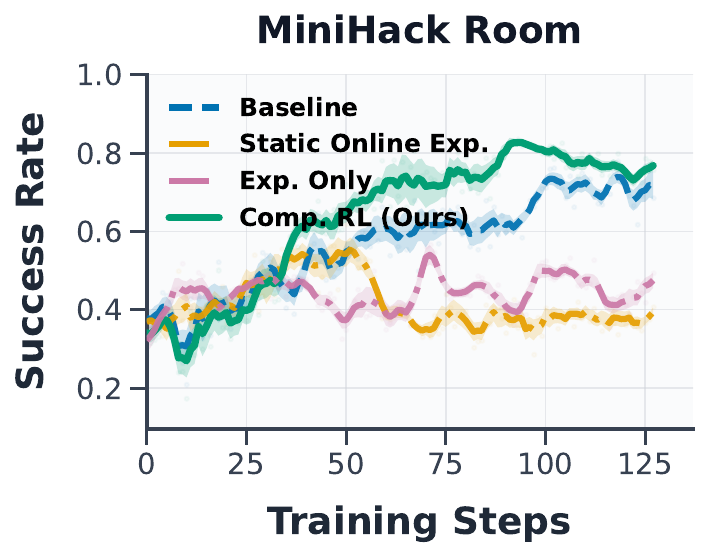}
        \vspace{-5mm}
      \end{subfigure}
      \begin{subfigure}[t]{0.24\textwidth}
        \centering
        \includegraphics[width=\linewidth]{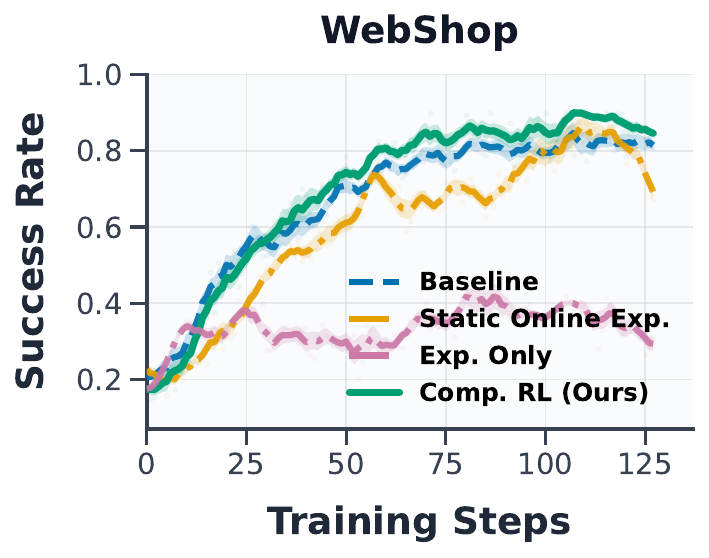}
        \vspace{-5mm}
      \end{subfigure}
      \begin{subfigure}[t]{0.24\textwidth}
        \centering
        \includegraphics[width=\linewidth]{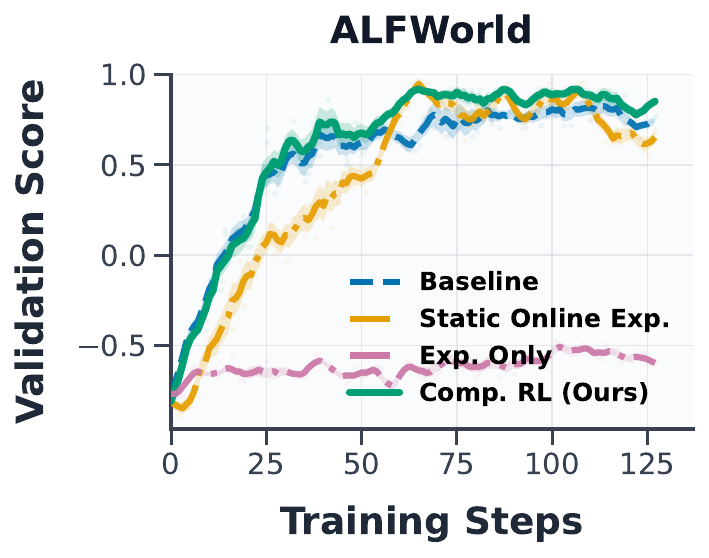}
        \vspace{-5mm}
      \end{subfigure}
      \vspace{-3mm}
        \caption{
        Multi-task training curves on overall and per-task performance.}
        \vspace{-5mm}
        \label{fig:main_result_multitask}
\end{figure}

\paragraph{Multi-Task Training}
\label{sec:multitask_training_exp}
Instead of training each task separately, we jointly train on MiniHack Room, ALFWorld, and WebShop to investigate whether \texttt{Complementary RL} 
can further benefit from cross-task experience distillation. We compare against three baselines that ablate the co-evolutionary design: 
\textbf{(1) Baseline:} actor training without any experience; 
\textbf{(2) Static Online Exp.:} $\pi_{\phi}$ dynamically maintains and constructs $\mathcal{M}$ during training but is not optimized, isolating the effect of extractor co-evolution; and \textbf{(3) Exp. Only:} $\pi_{\phi}$ is trained to maintain and refine $\mathcal{M}$, but the actor $\pi_{\theta}$ is held fixed, isolating the effect of actor co-evolution. Together, these baselines allow us to disentangle the mutual benefit of co-evolving both 
$\pi_{\theta}$ and $\pi_{\phi}$.
Table~\ref{tab:multitask_eval_perf} reports final evaluation performance, and 
Figure~\ref{fig:main_result_multitask} shows the training curves for each method. 
We also provide the number of actions per task throughout training in Appendix~\ref{appendix:multitask_num_actions}.
For methods that leverage experience, we evaluate under two settings: with and 
without retrieving from $\mathcal{M}$ at test time.

\begin{wrapfigure}{r}{0.6\textwidth}
    \begin{minipage}{0.6\textwidth}
        \centering  
        \vspace{-4mm}
        \captionof{table}{Multi-task evaluation performance. Methods with 
        (\textit{w/} exp.) retrieve experience from $\mathcal{M}$ at test time, 
        while (\textit{w/o} exp.) evaluates the actor $\pi_{\theta}$ alone.}
        \centering
        \label{tab:multitask_eval_perf}
            \resizebox{\textwidth}{!}{%
            \begin{tabular}{@{}lcccc@{}}
            \toprule
                                                         & Minihack Room & WebShop & ALFWorld & Avg. \\ \midrule
            Baseline                                     & 0.68          & 0.81    & 0.72     & 0.75 \\
            Static Online Exp. (eval \textit{w/} exp.)   & 0.41          & 0.67    & 0.69     & 0.59 \\
            Static Online Exp. (eval \textit{w/o} exp.) & 0.39          & 0.59    & 0.64     & 0.54 \\
            Exp. Only                                    & 0.49          & 0.37    & 0.13     & 0.33 \\\rowcolor{cyan!50}
            Comp. RL (eval \textit{w.} exp.)             & 0.78          & 0.87    & 0.82     & 0.82 \\\rowcolor{cyan!50}
            Comp. RL (eval \textit{w/o} exp.)            & 0.75          & 0.84    & 0.74     & 0.78 \\ \bottomrule
            \end{tabular}%
            }
        \vspace{-3mm}
    \end{minipage}
\end{wrapfigure}
The results reveal several key findings. First, integrating experience at test time consistently improves performance (e.g., $+5\%$ for both \textbf{Static Online Exp.} and \texttt{Complementary RL}), confirming the value of retrieved experience during inference. 
However, \textbf{Static Online Exp.} fails to surpass the baseline even with experience at test time (gap $>10\%$), and its training curves are dominated by the baseline across nearly all tasks. We attribute this to \textbf{distributional misalignment}: without parametric updates, the fixed extractor cannot adapt its experience maintenance strategy to the evolving actor, leading to noisy and inconsistent retrieval, particularly 
in the multi-task setting where cross-task experience contamination is observed.
In contrast, \texttt{Complementary RL} consistently outperforms the baseline 
both with and without experience at test time ($+7\%$ and $+2\%$ on average, respectively), demonstrating that co-evolutionary training internalizes useful experience into the actor itself. Finally, optimizing only the experience extractor (\textbf{Exp. Only}) yields marginal actor improvement, suggesting that experience quality alone is insufficient when the actor's base capability is limited~\citep{ouyang2025reasoningbank}.

\subsection{Analysis}
\label{sec:analysis}
\begin{figure}[t]
  \centering
      \begin{subfigure}[t]{0.24\textwidth}
        \centering
        \includegraphics[width=\linewidth]{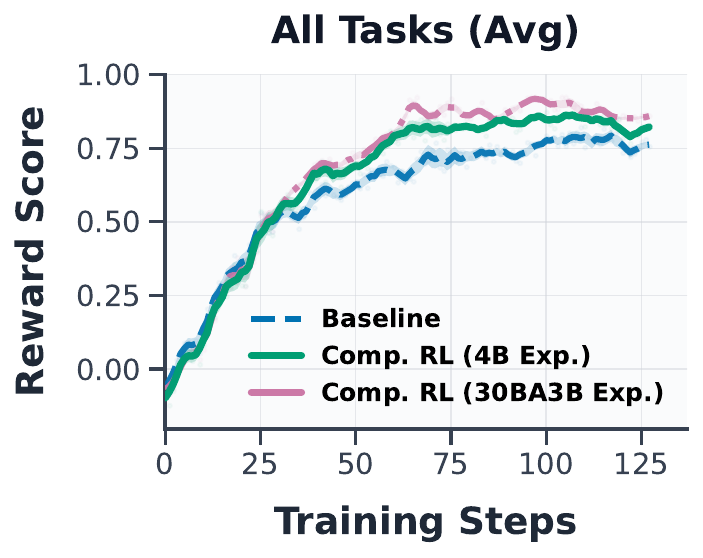}
        \vspace{-5mm}
        \caption{Extractor Capacity}
        \label{sub_fig:analysis_capacity}
      \end{subfigure}
      \begin{subfigure}[t]{0.24\textwidth}
        \centering
        \includegraphics[width=\linewidth]{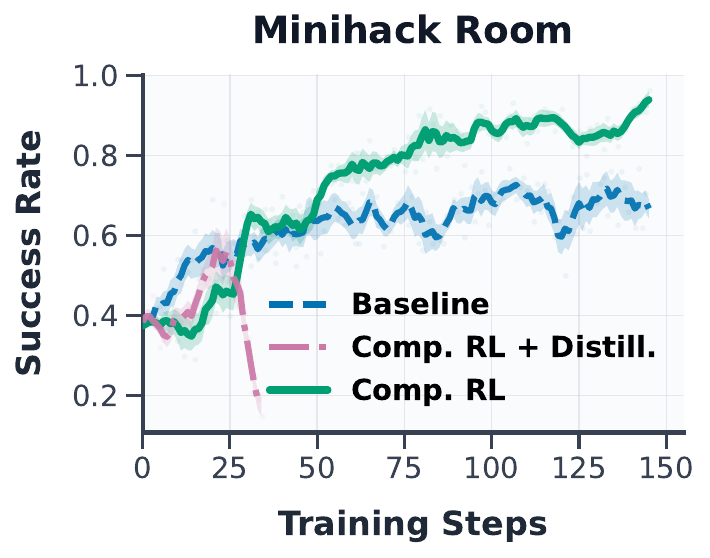}
        \vspace{-5mm}
        \caption{Comp. RL \textit{w/} Distill.}
        \label{sub_fig:analysis_distill}
      \end{subfigure}
      \begin{subfigure}[t]{0.24\textwidth}
        \centering
        \includegraphics[width=\linewidth]{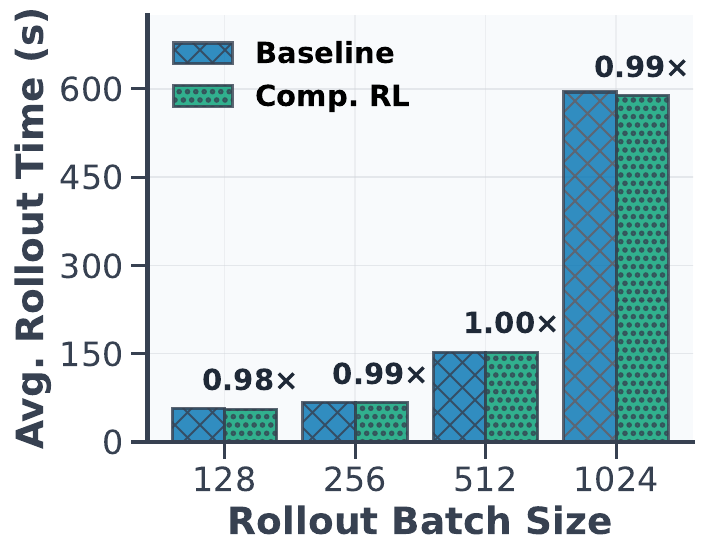}
        \vspace{-5mm}
        \caption{Rollout Time}
        \label{sub_fig:rollout_latency}
      \end{subfigure}
      \begin{subfigure}[t]{0.24\textwidth}
        \centering
        \includegraphics[width=\linewidth]{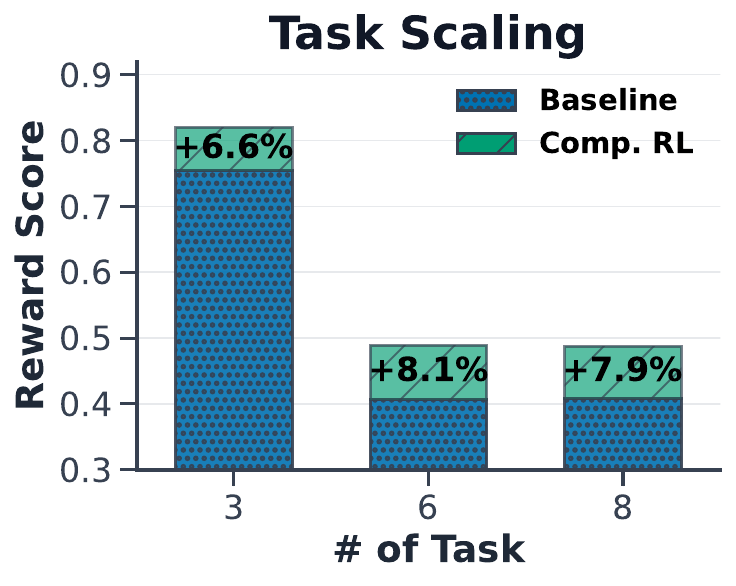}
        \vspace{-5mm}
        \caption{Task Scaling}
        \label{sub_fig:task_scaling_bar}
      \end{subfigure}
      \vspace{-2mm}
        \caption{Analysis of \texttt{Complementary RL} across different aspects of its design.}
      \vspace{-6mm}
      \label{fig:analysis_figure}
\end{figure}

\paragraph{Effect of Experience Extractor Capacity}
We investigate whether a stronger experience extractor $\pi_{\phi}$ can further amplify the benefits of \texttt{Complementary RL}. Specifically, we compare \texttt{Qwen3-30B-A3B-Instruct-2507} against the default \texttt{Qwen3-4B-Thinking-2507} as the experience extractor in multi-task training. 
As shown in Figure~\ref{sub_fig:analysis_capacity}, a larger 
experience extractor yields consistent improvement across tasks ($+5\%$ on 
average), suggesting that greater extractor capacity enables the extraction of more generalizable and informative experience, which in turn further benefits actor learning. Per-task results are provided in Appendix~\ref{appendix:per_task_capacity}.

\paragraph{Complementary RL with Self-Distillation}
Inspired by self-distillation~\citep{hubotter2026reinforcement}, we explore integrating self-distillation into \texttt{Complementary RL}.
For each trajectory in the experience-guided subgroup, we compare its score against the mean score of the experience-free subgroup; trajectories that exceed this threshold are collected into a self-distillation batch.
For each sample in this batch, we strip all experience-related context, including the retrieved experience at the first turn and all \texttt{search\_and\_ask} interactions, and supervise the actor $\pi_{\theta}$ via next-token prediction loss jointly with the RL objective.
This integration yields a dual benefit \texttt{Complementary RL} continues to optimize the actor through outcome-based rewards and evolving experience, while self-distillation additionally enables the actor to internalize successful experience-guided behaviors directly into its parameters, converting externally scaffolded reasoning into intrinsic capability.

However, results on MiniHack Room in Figure~\ref{sub_fig:analysis_distill} show that, while this integration initially improves upon \texttt{Complementary RL}, it collapses in later training.
We suspect this may stem from suboptimal hyperparameter choices, or alternatively, applying self-distillation at periodic intervals rather than every step may alleviate this issue.
Due to resource constraints, we leave additional investigation to future work.

\paragraph{Rollout Latency}
We run a series of experiments to evaluate whether our framework introduces additional latency to rollout collection during training.
We compare our framework against a baseline without experience integration and measure the average rollout collection time across different rollout batch sizes (i.e., varying numbers of parallel running environments).
Across all settings, we fix the number of parallel search workers and the batch processing query size as described in Appendix~\ref{appendix:training_configuration}.
As shown in Figure~\ref{sub_fig:rollout_latency}, our framework introduces no appreciable latency to rollout collection across all settings, remaining consistently on par with the baseline.
We also provide the detailed average search time per step during training in Figure~\ref{fig:average_search_time_wrt_batch_size}.

\paragraph{Task Scaling}
We next investigate whether \texttt{Complementary RL} continues to deliver benefits over the RL baseline without experience integration as the number of tasks scales up, more closely reflecting real-world industrial post-training settings where a broad mixture of tasks is used for RL.
To this end, in addition to the three-task mixture introduced in Section~\ref{sec:multitask_training_exp}, we further construct a six-task mixture by incorporating more challenging tasks; detailed configurations of the mixture are provided in Appendix~\ref{sec:task_mixture}.
The results are presented in Figure~\ref{sub_fig:task_scaling_bar}, which shows that \texttt{Complementary RL} consistently outperforms the baseline in both settings ($+6.6\%$ and $+8.1\%$ on the 3-task and 6-task mixtures, respectively), demonstrating that the performance gains of \texttt{Complementary RL} scale robustly with the number of tasks.
\section{Related Works}
\label{sec:related_works}
Leveraging accumulated experience to accelerate reinforcement learning has 
garnered significant attention for its potential to improve training efficiency~\citep{sutton2025, zhao2025absolute,zhai2025agentevolver}.
A direct approach is to store historical trajectories or workflows and 
retrieve them at inference time to improve performance on similar 
situations~\citep{moeini2025survey, deng2025rea, wang2024agentworkflowmemory,li2025memos}. 
However, such approaches cannot guarantee the quality or relevance of retrieved 
experience, potentially introducing noise that hinders learning.
To address this, one line of work introduces a dedicated experience extractor that 
dynamically constructs and maintains the experience bank in accordance with 
the agent's learning progress~\citep{xia2026skillrl, zhai2025agentevolver, 
zhang2026memrl}, while another line optimizes the retrieval process to ensure 
that high-quality and relevant experience is surfaced for agent 
improvement~\citep{zhang2026memrl, zhou2025memento}.
However, these works treat experience as a static resource, either maintaining fixed experience banks or employing non-adaptive extractors decoupled from the agent's evolving capabilities, which limits the full potential of the learning-from-experience paradigm. 
In contrast, \texttt{Complementary RL} co-evolves the agent and experience extractor, enabling dynamic and mutually beneficial adaptation throughout training.

Another key question is how to effectively utilize experience during RL 
training. 
The most straightforward approach treats experience as context, including it when computing policy gradients during RL 
optimization~\citep{li2025memos, salama2503meminsight, zhang2025learn, 
xia2026skillrl}.
However, this paradigm cannot guarantee improved performance 
when experience is absent at test time.
One line of work addresses this by decoupling rollout collection and policy optimization: experience is provided during rollout collection, while policy gradients are computed without experience in context, with the trust region adjusted accordingly~\citep{zhai2025agentevolver}.
Another line of work leverages experience to collect high-quality successful trajectories and optimizes the policy to reproduce them without experience in context~\citep{hubotter2026reinforcement, song2026expanding}.
In contrast, \texttt{Complementary RL} not only orchestrates the co-evolutionary training of both models, but also introduces experience-guided and experience-free rollout groups with separate advantage estimation for joint optimization under both conditions.

\section{Conclusion}
In this work, we present \texttt{Complementary RL}, a unified algorithm and infrastructure co-design framework that enables agents to effectively leverage and accumulate experience throughout the RL training process.
Rather than treating experience construction and management as a static component with a fixed extractor, we propose jointly training the policy actor and the experience extractor within an asynchronous dual-loop.
This co-evolutionary design ensures that the actor's growing capabilities continuously reshape what the extractor learns to distill, while the extractor's improving outputs in turn accelerate the actor's learning, each mutually and continuously shaping the other toward better performance.

\section{Acknowledgement}
We would like to thank Johan Obando-Ceron for the valuable discussions and feedback.

\clearpage
\bibliography{biblio}
\bibliographystyle{colm2024_conference}

\clearpage
\appendix
\newpage
\section{Additional Result}
\subsection{Action Efficiency Under Multi-Task Training}
\label{appendix:multitask_num_actions}
\begin{figure}[th]
  \centering
      \begin{subfigure}[t]{0.3\textwidth}
        \centering
        \includegraphics[width=\linewidth]{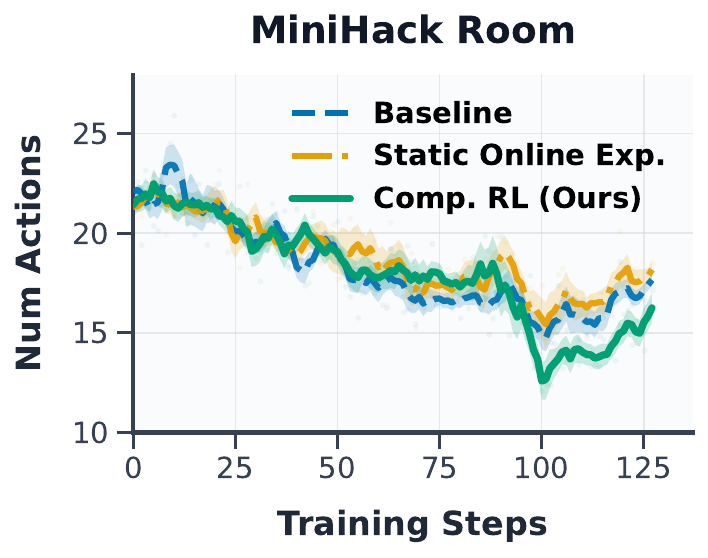}
        \vspace{-5mm}
      \end{subfigure}
      \begin{subfigure}[t]{0.3\textwidth}
        \centering
        \includegraphics[width=\linewidth]{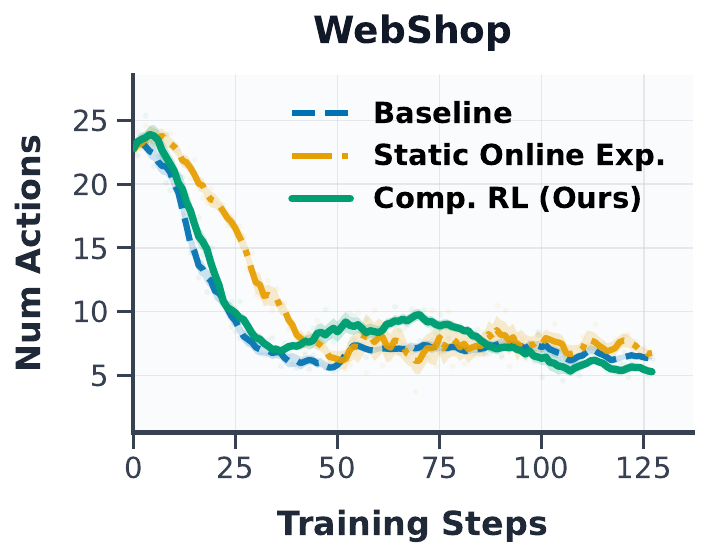}
        \vspace{-5mm}
      \end{subfigure}
      \begin{subfigure}[t]{0.3\textwidth}
        \centering
        \includegraphics[width=\linewidth]{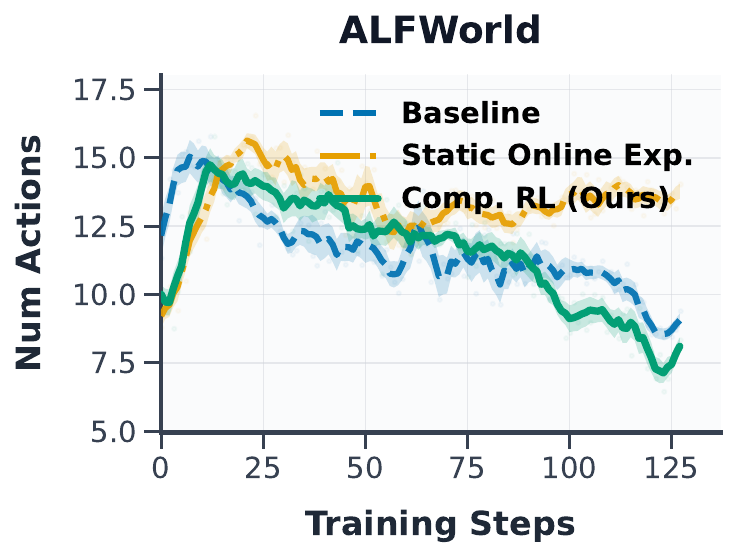}
        \vspace{-5mm}
      \end{subfigure}
      \vspace{-3mm}
        \caption{Average number of actions per task throughout multi-task training (corresponding to Figure~\ref{fig:main_result_multitask}).}
        \vspace{-5mm}
        \label{fig:num_actions_multitask}
\end{figure}

We further report the average number of actions per task during multi-task RL training in Figure~\ref{fig:num_actions_multitask}. The results consistently show that \texttt{Complementary RL} achieves superior action efficiency alongside 
higher success rates, further demonstrating the benefit of co-evolutionary experience in the multi-task setting.

\subsection{Per-Task Performance with Stronger Experience Extractor}
\label{appendix:per_task_capacity}
\begin{figure}[th]
  \centering
      \begin{subfigure}[t]{0.3\textwidth}
        \centering
        \includegraphics[width=\linewidth]{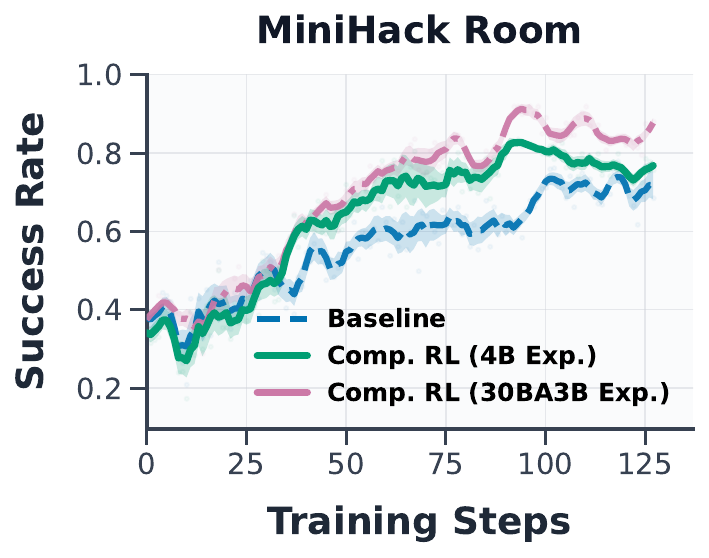}
        \vspace{-5mm}
      \end{subfigure}
      \begin{subfigure}[t]{0.3\textwidth}
        \centering
        \includegraphics[width=\linewidth]{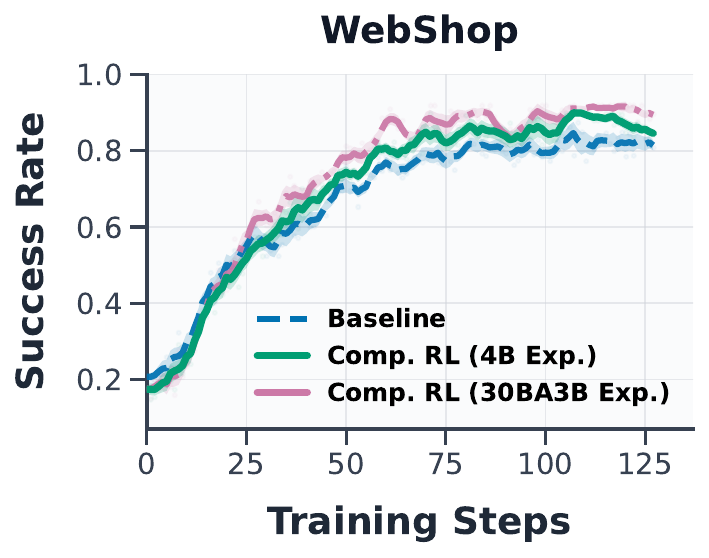}
        \vspace{-5mm}
      \end{subfigure}
      \begin{subfigure}[t]{0.3\textwidth}
        \centering
        \includegraphics[width=\linewidth]{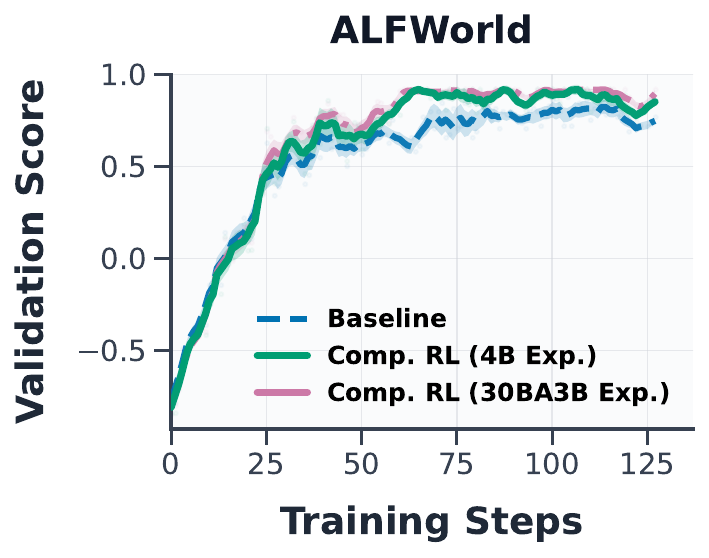}
        \vspace{-5mm}
      \end{subfigure}
      \vspace{-3mm}
        \caption{Per-task training dynamic (corresponding to Figure~\ref{sub_fig:analysis_capacity}).}
        \vspace{-5mm}
        \label{fig:per_task_perf_capacity}
\end{figure}

Figure~\ref{fig:per_task_perf_capacity} presents per-task performance in 
multi-task training across two experience extractor sizes (4B and 30B-A3B). 
Results show that a larger experience extractor consistently yields greater benefit across all tasks.

\subsection{Search Time Throught Training}
\begin{wrapfigure}{r}{0.3\textwidth}
    \begin{minipage}{0.3\textwidth}
        \centering  
        \vspace{-5mm}
        \centering
        \scalebox{1.00}
        {
            \begin{subfigure}[t]{\textwidth}
                \centering
                \includegraphics[width=\linewidth]{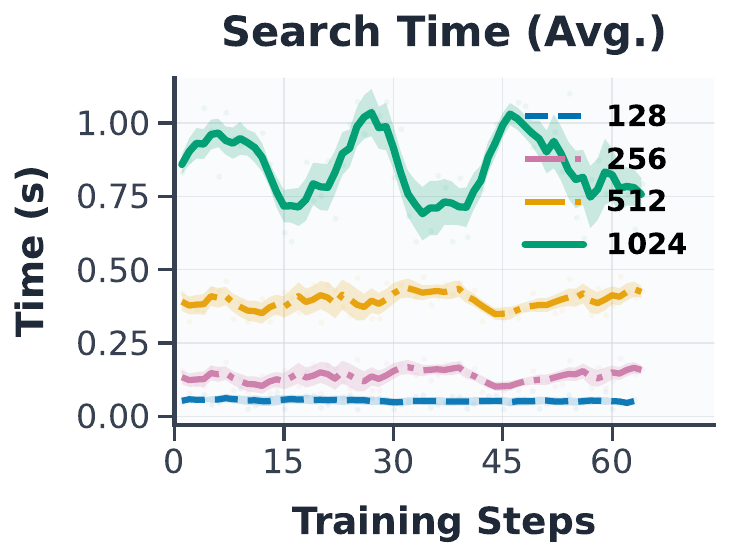}
            \end{subfigure}
        }
        \vspace{-7mm}
        \caption{Average search time per training step across different rollout batch sizes.}
        \label{fig:average_search_time_wrt_batch_size}
        \vspace{-9mm}
    \end{minipage}
\end{wrapfigure}
We further report the detailed average search time across all environments and training steps, corresponding to the experiment in Figure~\ref{sub_fig:rollout_latency}.
The results are presented in Figure~\ref{fig:average_search_time_wrt_batch_size}, which shows that although search time increases with larger rollout batch sizes, the maximum observed search time remains around 1 second, which is negligible.
We believe that by carefully tuning the query batch size $B$, the maximum waiting time $T_{\max}$, and the number of parallel search workers, the search time can be further reduced even in large rollout batch settings.

\subsection{Training Curves for Task Scaling Experiments}
\begin{figure}[th]
  \centering
      \begin{subfigure}[t]{0.45\textwidth}
        \centering
        \includegraphics[width=\linewidth]{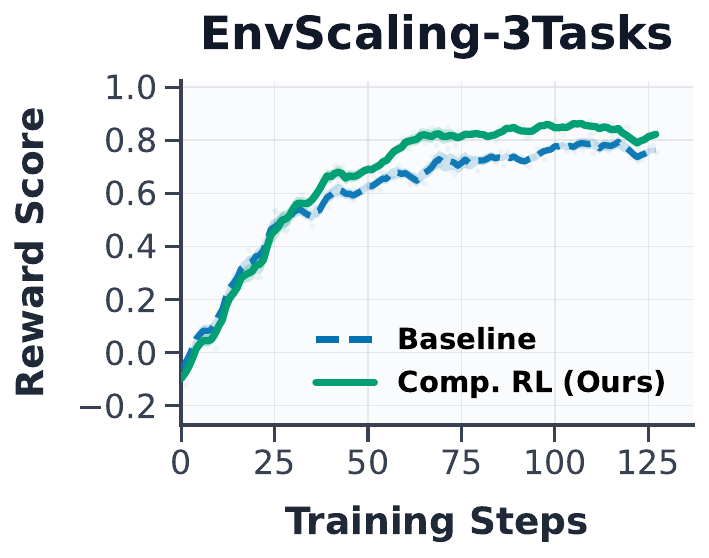}
        \vspace{-5mm}
      \end{subfigure}
      \begin{subfigure}[t]{0.45\textwidth}
        \centering
        \includegraphics[width=\linewidth]{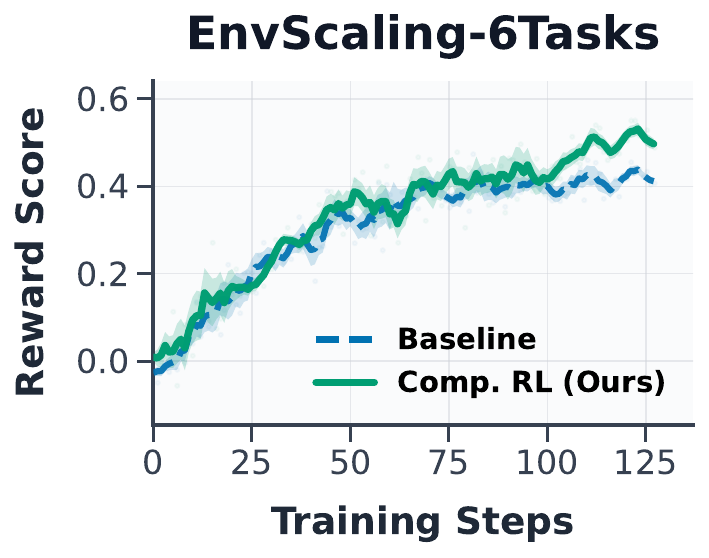}
        \vspace{-5mm}
      \end{subfigure}
      \vspace{-3mm}
      \caption{Training curves for \texttt{Complementary RL} and the baseline across different task mixture settings.}
        \vspace{-5mm}
        \label{fig:train_curve_task_mixture}
\end{figure}

We provide the training curves for the task scaling experiments introduced in Section~\ref{sec:analysis} in Figure~\ref{fig:train_curve_task_mixture}.

\section{Implementation Tricks}
\label{appendix:impl_tricks}
Training the experience extractor $\pi_{\phi}$ is highly unstable due to two compounding challenges.
First, the training is severely \textbf{off-policy}: since retrieval timing is uncontrolled, a distilled experience $m$ may be retrieved long after it was generated, introducing a large policy lag between the retrieving actor and the current $\pi_{\phi}$.
Second, when task descriptions exhibit low diversity, a single experience $m$ tends to be retrieved repeatedly across different training buffer steps, causing \textbf{data redundancy} and $\pi_{\phi}$ may be updated multiple times on the same experience $m$, severely undermining training stability.
To address these challenges, we introduce two stabilization techniques: \textbf{Retrieval Diversification} and \textbf{Training-Count-Aware Advantage Reweighting}.

\paragraph{Retrieval Diversification}
For each retrieval query $q$, instead of retrieving only the top-$K$ most relevant experiences, we oversample by drawing $N$ independent candidate sets, each of size $K$, yielding a total pool of $N \times K$ candidate experiences $\mathcal{C}(q) = \{m_1, m_2, \ldots, m_{N \times K}\}$.
We then re-rank the oversampled candidate pool $\mathcal{C}(q)$ according to a diversity-aware scoring function that penalizes frequently retrieved experiences:
\begin{equation}
    s(m) = s_{\text{rank}}(m) - \lambda \cdot \log(1 + c(m)) - \mathbb{1}[\text{recent}(m)],
\end{equation}
where $s_{\text{rank}}(m)$ is the base relevance rank score of experience $m$, $c(m)$ denotes its historical retrieval count, $\lambda$ is a penalty hyperparameter controlling retrieval diversity, and $\mathbb{1}[\text{recent}(m)]$ is an indicator that penalize experiences retrieved within a predefined recency window. The final top-$K$ experiences $\mathcal{R}(q)$ are selected as the highest-scoring entries under $s(m)$.
With this diversification strategy, $\pi_{\phi}$ is exposed to a broader and more varied set of experiences during training, mitigating the data redundancy issue and producing more diverse advantage signals for stable extractor optimization.

\paragraph{Training-Count-Aware Advantage Reweighting}
We observe that a single experience $m$ may be retrieved across multiple training buffer steps and optimized repeatedly, leading to overfitting and training instability for $\pi_{\phi}$.
To mitigate this, we reweight the advantage of each experience in the training buffer $\mathcal{B}_{\phi}$ according to both its cumulative training count and its recency of optimization.
Specifically, after computing the advantage for each sample in $\mathcal{B}_{\phi}$ according to Section~\ref{sub_sec:comp_rl}, we apply a per-experience weight $w(m)$ defined as:
\begin{equation}
    w(m) = \begin{cases}
        0 & \text{if } (t - t_{\text{last}}) < \delta, \\
        (1 + c_{\text{train}}(m))^{-\alpha} & \text{otherwise},
    \end{cases}
\end{equation}
where $t$ is the current global training step, $t_{\text{last}}$ is the most recent step at which $m$ was trained, $\delta$ is a cooldown window that suppresses gradient updates from experiences optimized too recently, $c_{\text{train}}(m)$ is the cumulative number of times $m$ has been trained on, and $\alpha \geq 0$ is a decay exponent controlling how aggressively the advantage is discounted as $m$ accumulates training counts. 
Together, the cooldown mechanism prevents repeated optimization within a short window, while the count-based decay progressively down-weights overused experiences, yielding more stable and balanced training of $\pi_{\phi}$.

\subsection{Actor Critic}
During training of \texttt{Complementary RL}, we observe that retrieved experiences can sometimes confuse rather than benefit the actor, particularly in the early stages of training.
Upon closer inspection, we identify two failure modes: 
(1) \textbf{Experience Staleness}: when the actor has already mastered a given task, the retrieved experience may be overly conservative or even incorrect relative to the actor's current capability, thereby degrading performance rather than improving it;
(2) \textbf{Experience Imprecision}: when the actor's success rate is low, retrieved experiences are often directionally helpful but may require adaptation to the task at hand, as they are not always precisely aligned with the current context.
To address the above failure modes, we propose \texttt{Actor-Critic}, which introduces explicit communication between the policy actor $\pi_{\theta}$ and the experience extractor $\pi_{\phi}$.

Specifically, prior to launching the main dual training loop, we run $\pi_{\theta}$ for $T_{\text{warm}}$ warm-up iterations on the current task to estimate its initial average success rate $\bar{r}_{\theta}$.
Once training begins, after each retrieval of experience $m$ for a given task query $q$, we prompt the actor $\pi_{\theta}$ to reflect on the retrieved experience in light of both the current task and its accumulated success rate $\bar{r}_{\theta}(q)$.
Based on this reflection, the actor produces one of three critic actions:
\begin{itemize}
    \item \texttt{accept}: the experience $m$ is accepted as-is, receiving a critic score $s_c(m) = 1$;
    \item \texttt{refine}: the experience $m$ is refined using the actor's own knowledge, receiving a critic score $s_c(m) = 0.5$;
    \item \texttt{reject}: the experience $m$ is discarded, receiving a critic score $s_c(m) = 0$.
\end{itemize}
This mechanism allows the actor to selectively consume experience commensurate with its current capability. Furthermore, the critic score $s_c(m)$ is combined with the task completion reward $r(m)$---the outcome obtained when using experience $m$ to solve the task---to form an enriched learning signal for the experience extractor $\pi_{\phi}$:
\begin{equation}
    \tilde{r}(m) = s_c(m) + r(m).
\end{equation}

\begin{figure}[t]
  \centering
      \begin{subfigure}[t]{0.45\textwidth}
        \centering
        \includegraphics[width=\linewidth]{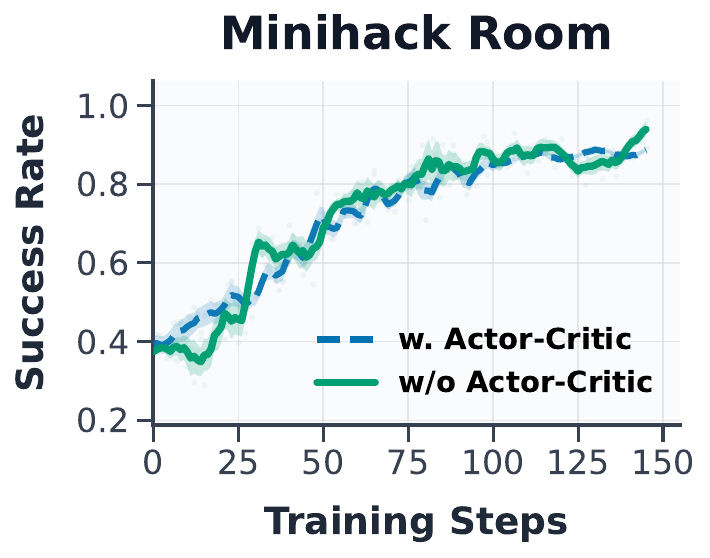}
        \vspace{-5mm}
        \caption{Success Rate}
        \label{sub_fig:actor_critic_performance}
      \end{subfigure}
      \begin{subfigure}[t]{0.45\textwidth}
        \centering
        \includegraphics[width=\linewidth]{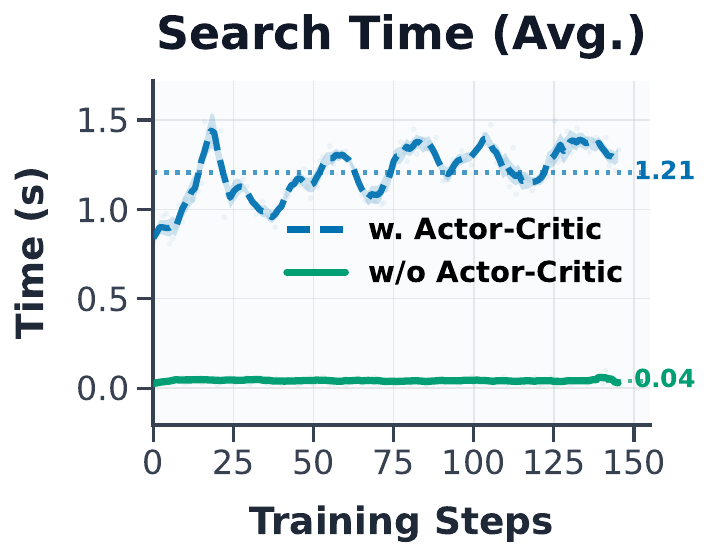}
        \vspace{-5mm}
        \caption{Retrieval Speed}
        \label{sub_fig:actor_critic_speed}
      \end{subfigure}
      \vspace{-3mm}
        \caption{Analysis of incorporating \texttt{Actor-Critic} into \texttt{Complementary RL}.}
        \vspace{-5mm}
        \label{fig:actor_critic_analysis}
\end{figure}

As shown in Figure~\ref{sub_fig:actor_critic_performance}, \texttt{Actor-Critic} yields improved success rates, particularly in the early stages of training on MiniHack Room.
However, since the actor must produce a critic decision before each environment interaction, rollout collection is blocked pending the critic result, incurring non-trivial latency overhead (Figure~\ref{sub_fig:actor_critic_speed}).
Therefore, we do not adopt \texttt{Actor-Critic} as a default component in our main experiments, but recommend it as a beneficial addition in scenarios where final performance is prioritized over training throughput.

\subsection{Lessons Learned}
\label{appendix:learned_lessons}
\paragraph{Separate Model Parameters for Actor and Extractor}
In early attempts, we served a single set of parameters shared between the training and inference engines, using the same weights for both the policy actor $\pi_{\theta}$ and the experience extractor $\pi_{\phi}$.
This design optimizes a single model under two distinct objectives simultaneously (Equation~\ref{eq:grpo_split} and~\ref{eq:cispo}).
However, since the two optimization objectives impose possible conflicting gradient directions, we were unable to guarantee stable training despite extensive tuning efforts.
We ultimately resolved this by maintaining separate parameter suites for the actor $\pi_{\theta}$ and the experience extractor $\pi_{\phi}$, which decouples the two optimization objectives and yields stable training.

\paragraph{Direct Reward over Relative Reward for Experience}
In early attempts, rather than using the actor's task completion reward as the direct reward signal $r(m)$ for experience $m$, we explored a relative reward strategy.
Specifically, we first computed the average reward of the experience-free subgroup as a baseline, and then assigned each sample in the experience-guided subgroup a reward proportional to its improvement over this baseline.
However, empirical comparison revealed that this relative reward strategy consistently underperforms direct reward assignment, and we therefore adopt the latter in \texttt{Complementary RL}.

\paragraph{Auxiliary Perplexity Reduction Reward}
For challenging tasks, a retrieved experience may be instructive yet insufficient to directly yield task success.
We therefore explored augmenting the reward signal for $\pi_{\phi}$ with a perplexity reduction bonus, motivated by the intuition that a genuinely helpful experience should increase the actor's confidence, and thus reduce its entropy, when processing the task at hand.

Concretely, at the start of each task, we compute the actor's entropy $\mathcal{H}(\pi_{\theta} \mid q)$ over the task query $q$ without any retrieved experience, and then re-compute the entropy $\mathcal{H}(\pi_{\theta} \mid q, m)$ after injecting the retrieved experience $m$ into the system message.
The entropy reduction $\Delta\mathcal{H}(m) = \mathcal{H}(\pi_{\theta} \mid q) - \mathcal{H}(\pi_{\theta} \mid q, m)$ is then used as an auxiliary reward bonus for $\pi_{\phi}$.
We evaluated five normalization strategies for computing this bonus:
\begin{itemize}
    \item \textbf{Relative}: scale-invariant percentage reduction, $b = w \cdot \Delta\mathcal{H}(m) / \mathcal{H}(\pi_{\theta} \mid q)$, clipped to a predefined range;
    \item \textbf{Tanh}: smooth non-linear scaling bounded to $[-1, 1]$, $b = w \cdot \tanh(\Delta\mathcal{H}(m))$;
    \item \textbf{Sigmoid}: temperature-scaled sigmoid bounded to $[0, 1]$ and re-centered, $b = w \cdot (2\sigma(\Delta\mathcal{H}(m) / \tau) - 1)$;
    \item \textbf{Asymmetric Clipping}: asymmetrically clips negative and positive gains to encourage exploration without heavy penalty;
    \item \textbf{Log-Space}: sign-preserving logarithmic compression, $b = w \cdot \text{sgn}(\Delta\mathcal{H}(m)) \cdot \log(1 + |\Delta\mathcal{H}(m)|)$, suitable when entropy magnitudes vary widely.
\end{itemize}
However, none of these strategies yielded a consistent improvement in practice, and we therefore exclude this auxiliary reward from \texttt{Complementary RL}.

\section{Implementation Details}
\label{appendix:impl_detail}
\subsection{Environment Description}
\label{appendix:env_description}
During RL training, we implement each environment following the protocol of GEM~\citep{liu2026gemgymagenticllms}, with the the SWE task is additionally executed using ROCK\footnote{\url{https://github.com/alibaba/ROCK}}.
All environments adopt a binary reward scheme, assigning $r = 1$ upon task success and $r = 0$ upon failure, with the exception of ALFWorld, which assigns $r = 1$ upon success and $r = -1$ upon failure.
In the following, we provide a brief introduction to each environment and our corresponding implementation details.

\paragraph{MiniHack}
MiniHack~\citep{samvelyan2021minihack} is a collection of game environments in which an agent explores a world under a fog-of-war observation model, meaning the agent can only observe its immediately surrounding grid cells.
The goal of the agent is to reach a target destination by avoiding traps and obstacles, using tools to cross rivers or lava, or defeating monsters.
We adapt MiniHack for LLM-based agents by representing each entity---including items, traps, monsters, and the agent itself---as a text symbol following the NetHack convention\footnote{\url{https://nethackwiki.com/wiki/Main_Page}} (e.g., \texttt{@} represents the agent's position, \texttt{>} represents the goal position).
At each timestep, the agent is provided with the current observable grid layout along with a legend of symbol meanings, and is asked to decide the next action.
The action space typically consists of directional movements, with additional task-specific actions such as \texttt{pick\_up} or \texttt{apply} in more complex environments.

In this work, we evaluate on the following MiniHack environments of increasing difficulty:
\begin{itemize}
    \item \textbf{MiniHack Room}\footnote{\url{https://minihack.readthedocs.io/en/latest/envs/navigation/room.html}}: The agent navigates a dark room, avoiding traps, obstacles, and monsters to reach the goal. We use \textit{MiniHack-Room-Ultimate-5x5-v0}. The action space consists solely of directional actions (e.g., north, south, east, west).

    \item \textbf{MiniHack Maze}\footnote{\url{https://minihack.readthedocs.io/en/latest/envs/navigation/mazewalk.html}}: A more challenging environment in which the agent must navigate a maze to reach the goal. We use \textit{MiniHack-MazeWalk-9x9-v0}. The action space also consists of directional actions.

    \item \textbf{MiniHack KeyRoom}\footnote{\url{https://minihack.readthedocs.io/en/latest/envs/navigation/keyroom.html}}: The agent must first locate a key, find a locked door, open the door with the key, and finally reach the goal position. We use \textit{MiniHack-KeyRoom-Dark-S5-v0}. This environment includes additional actions beyond directional movement, such as \texttt{pick\_up} and \texttt{apply}.

    \item \textbf{MiniHack River\footnote{\url{https://minihack.readthedocs.io/en/latest/envs/navigation/river.html}}}: The agent must first push a boulder into the river, then cross it, and finally reach the goal. We use \textit{MiniHack-River-Narrow-v0}.
\end{itemize}

\paragraph{WebShop}
WebShop~\citep{yao2023webshopscalablerealworldweb} is a benchmark that simulates web-based shopping, in which agents navigate a realistic web interface to find and purchase products matching user specifications.
At each timestep, the agent receives a product specification and must choose between two types of actions: issuing a text search query (e.g., \texttt{search[red shoes]}) or clicking a text button (e.g., \texttt{choose[Size 9]}).
The environment returns an observation after each action, and the agent continues until the target product is successfully purchased or the episode terminates.
In our implementation, we adopt the small variant configuration, restricting the searchable product catalog to 1,000 items, with goals sampled from the instruction pool via weighted sampling based on attribute frequency.

\paragraph{ALFWorld}
ALFWorld~\citep{shridhar2021alfworldaligningtextembodied} is a text-based interactive environment aligned with the ALFRED embodied AI benchmark~\citep{shridhar2020alfredbenchmarkinterpretinggrounded}, in which agents complete household tasks by navigating rooms and interacting with objects through natural language commands.
Each task presents the agent with a high-level goal (e.g., \texttt{put a heated plate in the fridge}), and at each timestep, the agent receives a textual observation describing the objects visible in the current room and must issue a natural language action (e.g., \texttt{go to countertop 1}, \texttt{pick up plate}).
The episode terminates upon successful task completion or when the maximum number of steps is reached.
In our implementation, we train on 1,466 task instances from ALFWorld and hold out 134 instances for evaluation.

\paragraph{SWE-Bench}
SWE-Bench~\citep{jimenez2024swebenchlanguagemodelsresolve} is a real-world software engineering benchmark in which an agent must resolve GitHub issues by modifying the relevant portions of a codebase such that all provided unit tests pass successfully.
For each task instance, the agent receives a GitHub issue description and interacts with the codebase through three tools: \texttt{execute\_bash} for executing shell commands, \texttt{str\_replace\_editor} for viewing and editing source files, and \texttt{submit} for submitting the final patch.
The environment returns the tool execution result as a textual observation after each action.

In our experiments, we utilize SWE-Bench-Verified for training.
However, since many tasks in the full dataset are too challenging for smaller models, naively training \texttt{Qwen3-4B-Instruct-2507} on the complete dataset yields unstable and ineffective learning.
To address this, we perform a preliminary \texttt{pass@16} evaluation using \texttt{Qwen3-4B-Instruct-2507} and retain only tasks with a success rate in the range $(0, 80\%)$, filtering out both trivially easy and prohibitively difficult instances.
This yields a curated training set of 124 tasks, and we report the final success rate evaluated throughout training.

\paragraph{Sokoban}
Sokoban is a classic text-based puzzle game in which an agent must navigate a grid and push boxes onto designated target positions while avoiding walls.
The task requires multi-step planning and spatial reasoning, as boxes can only be pushed and an incorrectly pushed box may render the puzzle unsolvable.
We represent the walls, boxes, targets, agent, and empty positions using structured text symbols such as \texttt{W}, \texttt{A}, \texttt{C}, \texttt{@}, and \texttt{.}, respectively.
We configure each episode as a $6\times 6$ room with two boxes and two corresponding target positions, yielding a challenging combinatorial search space for the agent.

\subsection{Training Configuration}
\label{appendix:training_configuration}
We implement \texttt{Complementary RL} within the ROLL\footnote{\url{https://github.com/alibaba/ROLL}} framework, using Megatron as the training engine and vLLM as the inference engine across all experiments.
We do not apply KL regularization for either $\pi_{\theta}$ or $\pi_{\phi}$, and adopt the AdamW optimizer with a constant learning rate of $1\times 10^{-6}$ throughout.
Unless otherwise specified, we run 4 parallel search workers and 4 parallel embedding workers in our framework.
We use \texttt{Qwen3-Embedding-0.6B}\footnote{\url{https://huggingface.co/Qwen/Qwen3-Embedding-0.6B}} as the embedding model, served via vLLM.
The query batch size $B$ is set to 16, and the maximum waiting time $t_{\max}$ is set to 0.001 seconds.
The training buffer size $|\mathcal{B}_{\phi}|$ for $\pi_{\phi}$ is set to 64, and the periodic merge interval is set to 5 steps.
The importance sampling clip thresholds $\epsilon_{\text{low}}^{\text{IS}}$ and $\epsilon_{\text{high}}^{\text{IS}}$ in Equation~\ref{eq:cispo} are both set to 0.1.
In the following, we describe the specific implementation details for each experimental group.

\paragraph{Configuration of Experiments in Figure~\ref{fig:method_motivation}}
We run each experiment with a total rollout batch size of 128, a group size of $K = 8$, and a clip ratio of $\epsilon = 0.2$.
Each experiment runs for 145 steps with a micro-batch size of 64 for $\pi_{\theta}$.
We set the maximum number of interaction turns to 30, the maximum output tokens per step to 4,096, the maximum sequence length for $\pi_{\theta}$ to 32,768 tokens, and the maximum sequence length for $\pi_{\phi}$ to 65,536 tokens.
\begin{itemize}
    \item \textbf{Offline Exp.} We run \texttt{Qwen2.5-7B-Instruct} offline to interact with MiniHack Room for a maximum of 30 interaction turns. The resulting trajectories are then routed to \texttt{Qwen3-30B-A3B-Instruct-2507} for experience distillation, followed by the same merging and semantic-similarity-based deduplication pipeline used in our main experiments to construct a high-quality offline experience bank.

    \item \textbf{Static Online Exp.} This variant follows the same setup as \texttt{Complementary RL}, except that the experience extractor $\pi_{\phi}$ is not optimized during training. All other components remain active, including subgroup separation, query diversification, and \texttt{search\_and\_ask}.
\end{itemize}

\paragraph{Configuration of Single-Task Training}
Unless otherwise noted, all other settings follow the general configuration described above.
\begin{itemize}
    \item \textbf{WebShop}: We use a rollout batch size of 64, a group size of $K = 8$, and a micro-batch size of 16. The number of warmup steps for $\pi_{\theta}$ is set to 10, the maximum number of training steps is 256, and the maximum sequence length for $\pi_{\theta}$ is 16,384 tokens.

    \item \textbf{ALFWorld}: We use a rollout batch size of 128, a group size of $K = 8$, and a micro-batch size of 32. The maximum number of training steps is 128, the maximum sequence length for $\pi_{\theta}$ is 16,384 tokens, the maximum number of interaction turns is 40, and the maximum output tokens per step is 2,048.
\end{itemize}

\paragraph{Configuration of Multi-Task Training}
We run all experiments with a total rollout batch size of 384, with each task contributing a batch size of 128, a group size of $K = 8$, and a micro-batch size of 96.
Training runs for 128 steps, with a maximum sequence length of 32,768 tokens, a maximum interaction count of 30, and a maximum output tokens per step of 4,096.
All other settings follow the general configuration described above.

\subsection{Task Mixture}
\label{sec:task_mixture}
\paragraph{3-Tasks:} Minihack Room, Webshop, and ALFWorld.
\paragraph{6-Tasks:} Minihack Room, Minihack Maze, Minihack KeyRoom, Sokoban, Webshop, and ALFWorld.

\section{Illustration}
Here, we provide representative examples of distilled experience in our experiments.

\paragraph{Single-Task Experience}
We present representative distilled experience entries from single-task training for MiniHack (Table~\ref{tab:minihack_exp}), WebShop (Table~\ref{tab:webshop_exp}), ALFWorld (Table~\ref{tab:alfworld_exp}), and SWE-Bench (Table~\ref{tab:swe_exp}).

\paragraph{Multi-Task Experience}
We also find that the experience extractor is capable of distilling universal experience transferable across tasks, for which we show representative examples in Table~\ref{tab:multitask_exp}.

\begin{table}[t!]
    \centering
    \begin{minipage}{0.99\textwidth} 
    \centering
    \caption{Minihack distilled experience.}
    \begin{tcolorbox} 
        \centering
        \small
        \hspace{-6mm}
        \begin{tabular}{p{0.99\textwidth}}

            \begin{minipage}{0.99\textwidth}\vspace{0mm}
                \vspace{2mm}
                When an agent observes a visible exit (e.g., staircase down symbol '\texttt{>}') in the immediate field of view and the direct path to it is unobstructed by hazards or threats (e.g., no monsters, traps, or dead ends), prioritize moving toward that exit immediately. This strategy works best when the exit is in a cardinal direction with clear visibility and no blocking obstacles.

                \textbf{Action Sequence:}
                1. Identify visible exits (e.g., '\texttt{>}', 'door' symbols) within the current dungeon view.
                2. For each visible exit, verify the path to it has no immediate threats or hazards (e.g., jackals, traps) in the adjacent cells.
                3. Move in the direction of the nearest visible exit with an unobstructed path (e.g., north if the staircase is directly north).
                
                \textbf{Decision Logic:}
                - If a visible exit path is unobstructed, move toward it.
                - If threats are present but do not block the exit path, do not deviate toward threats—focus on moving toward the exit first.
                - If the environment offers a temporary benefit (e.g., "full moon" implying enhanced visibility or reduced aggression), prioritize the direct path to the exit over waiting for the benefit, unless the benefit explicitly reduces threat levels (e.g., "reduced monster aggression").
                
                \textbf{Failure Prevention:}
                - Do NOT move toward adjacent threats (e.g., jackals) if they do not block the exit path, as this increases exposure to potential hazards.
                - Do NOT delay action due to environmental effects (e.g., full moon) when a clear path exists, as the benefit may be passive and not directly actionable (e.g., no explicit effect on movement or threat reduction).
                - Do NOT assume environmental effects (e.g., full moon) provide direct benefits unless the environment explicitly states their impact (e.g., "reduced monster aggression").
                
                \textbf{Generalization:} This principle applies to any grid-based navigation task with visually distinguishable exits and localized threats (e.g., mazes, dungeons, code navigation). It ensures agents focus on immediate, actionable progress toward known goals without unnecessary delays caused by environmental conditions that lack explicit, actionable benefits.

            \end{minipage}
        \end{tabular}
    \end{tcolorbox}
    \label{tab:minihack_exp}
    \end{minipage}
\end{table}

\begin{table}[th!]
    \centering
    \begin{minipage}{0.99\textwidth} 
    \centering
    \caption{WebShop distilled experience.}
    \begin{tcolorbox} 
        \centering
        \small
        \hspace{-6mm}
        \begin{tabular}{p{0.99\textwidth}}

            \begin{minipage}{0.99\textwidth}\vspace{0mm}
                \vspace{2mm}
                When searching for products with specific color variants (e.g., 'type 3-camel') as a critical constraint, **first** filter the search results by this color variant to isolate relevant products, then verify the remaining products meet all other requirements (price, features, installation specs) before purchase.
                
                \textbf{1. **Initial search**:} Use broad keywords matching all non-color requirements (e.g., "height adjustable high density easy install easy assemble home office chairs price \texttt{<} 130").
                \textbf{2. **Color filter**:} From the search results, explicitly set the color to the required variant (e.g., `click[type 3-camel]`) to narrow results.
                \textbf{3. **Verification**:} Check the filtered list for products that satisfy the non-color constraints (e.g., price \texttt{<} \$130, features like "easy install").
                \textbf{4. **Proceed**:} If a product matches, select and purchase; if not, refine the search or adjust filters.
                
                \textbf{**Decision Logic**}:
                - *If* the initial search yields products with multiple color options and the task specifies a **particular** color variant, *then* apply the color filter to exclude irrelevant products.
                - *If* no products match after filtering, *then* re-run the search with adjusted keywords (e.g., broader color terms) or expand the search scope.
                
                \textbf{**Failure Prevention**}:
                - Do **not** skip color filtering when the task explicitly requires a specific color variant.
                - Do **not** assume products with similar names or descriptions match the exact color requirement.
                - **Why**: Color variants are often mislabeled or not clearly specified in product listings; filtering first ensures compliance with the task’s color constraint.
                
                \textbf{**Generalization**}: This strategy applies to any e-commerce environment where color variants are a critical requirement (e.g., furniture, apparel, electronics). It is invariant to the specific product category, price thresholds, or feature lists.
            \end{minipage}
        \end{tabular}
    \end{tcolorbox}
    \label{tab:webshop_exp}
    \end{minipage}
\end{table}

\begin{table}[th!]
    \centering
    \begin{minipage}{0.99\textwidth} 
    \centering
    \caption{ALFWorld distilled experience.}
    \begin{tcolorbox} 
        \centering
        \small
        \hspace{-6mm}
        \begin{tabular}{p{0.99\textwidth}}

            \begin{minipage}{0.99\textwidth}\vspace{0mm}
                \vspace{2mm}
                \textbf{1. **Situational Context**}: When a task requires locating a specific object (e.g., food items, tools) in a household environment and initial checks fail.
                
                \textbf{2. **Action Sequence**}:
                   a. Check all countertops (in numerical order from 1 to N) for the object. Countertops are common storage areas for frequently used items like food.
                   b. If the object isn't found on countertops, check the fridge (after opening it if closed) for perishables or stored food.
                   c. If still not found, check sinks (for washed items), cabinets (for stored items), and shelves (for items kept at eye level) in a logical order.
                
                \textbf{3. **Decision Logic**}:
                   - If the object is a food item (e.g., fruit, vegetables), prioritize countertops first as they are common for fresh items.
                   - If the object is a tool or consumable that's typically stored in containers (e.g., spices, packaged goods), check the fridge after countertops.
                   - If the object is small and likely hidden (e.g., a key, battery), adjust the order to check high shelves, drawers, or enclosed spaces first.
                
                \textbf{4. **Failure Prevention**}:
                   - Do not assume the object is in a specific location without checking (e.g., the agent didn't assume the apple was in the fridge despite checking it).
                   - Do not skip checking countertops when searching for food items (e.g., the agent first checked countertop 1 before the fridge).
                   - Avoid redundant searches by tracking which locations have been checked (e.g., noting "countertop 1 checked" to prevent re-checking).
                
                \textbf{5. **Generalization**}: This strategy applies universally to household environments where objects are stored in common locations. It works across variations in object types (e.g., food, tools, household items) and environments (e.g., different house layouts). It avoids coordinates (e.g., "countertop 2") by using relative terms like "numerical order" and "logical order" based on typical household practices.
            \end{minipage}
        \end{tabular}
    \end{tcolorbox}
    \label{tab:alfworld_exp}
    \end{minipage}
\end{table}

\begin{table}[th!]
    \centering
    \begin{minipage}{0.99\textwidth} 
    \centering
    \caption{SWE-Bench distilled experience.}
    \begin{tcolorbox} 
        \centering
        \small
        \hspace{-6mm}
        \begin{tabular}{p{0.99\textwidth}}

            \begin{minipage}{0.99\textwidth}\vspace{0mm}
                \vspace{2mm}
                When attempting to make precise code edits using string replacement tools (like the \texttt{str\_replace\_editor} function), follow this systematic approach to ensure success:
                
                \textbf{1. **Verify the actual code content**}:
                   - IF you plan to perform a string replacement:
                     - EXECUTE a `view` command with an appropriate line range to get the exact code
                     - Use the output to determine the precise string pattern matching the target code
                
               \textbf{2. **Extract the exact replacement pattern**}:
                   - IF the string you want to replace is multi-line:
                     - Capture the exact output from the `view` command with appropriate line ranges
                     - Include all whitespace (indentation, spaces, line breaks) in your replacement pattern
                   - IF the target pattern has variable whitespace (e.g., different indents):
                     - Use the most common or canonical whitespace pattern from your `view` output
                     - Consider adding a fallback for variations
                
                \textbf{3. **Handle failure cases**}:
                   - IF the tool reports no occurrences found:
                     a. Re-examine the `view` output for subtle differences
                     b. Check if whitespace differences exist (e.g., extra spaces, tabs, newlines)
                     c. Try matching with a slightly different pattern that accounts for common whitespace variations
                   - IF the code is in a version control system with line endings:
                     a. Normalize line endings before attempting the replacement
                     b. Ensure the replacement pattern matches the line endings (e.g., Unix vs. Windows)
                
                \textbf{4. **Alternative approach**}:
                   - IF precise string replacement fails repeatedly:
                     - Create a temporary file with the fix using the `\texttt{str\_replace}` tool
                     - Save the fixed file to a temporary location
                     - Then replace the original file using an atomic operation
                     - This avoids multiple failed attempts
                
                \textbf{5. **Verification**}:
                   - AFTER making a replacement:
                     a. Run `view` on the modified code to confirm the change
                     b. Test the code functionality (if possible) to ensure the fix worked
                     c. Check for potential side effects like unintended changes to other parts of the code
                
                This strategy ensures reliable code edits in environments with string-based editing tools by emphasizing precision in pattern matching and verification through explicit code validation.
                
                \textbf{Special considerations for Python and type systems:}
                - When working with type-related code (especially with TypeVar, generics, etc.):
                  - Be mindful that TypeVar objects are not iterable
                  - Always verify if a parameter is a tuple before iterating over it
                  - Add clear comments to explain non-iterable behavior for type safety
            \end{minipage}
        \end{tabular}
    \end{tcolorbox}
    \label{tab:swe_exp}
    \end{minipage}
\end{table}

\begin{table}[th!]
    \centering
    \begin{minipage}{0.99\textwidth} 
    \centering
    \caption{Universal distilled experience from multi-task training.}
    \begin{tcolorbox} 
        \centering
        \small
        \hspace{-6mm}
        \begin{tabular}{p{0.99\textwidth}}

            \begin{minipage}{0.99\textwidth}\vspace{0mm}
                \vspace{2mm}
                \textbf{1. Situational Context}: This protocol governs how an agent detects stagnation, 
                breaks out of unproductive loops, and escalates to higher-level problem-solving when 
                direct attempts repeatedly fail.
                
                \textbf{2. When to Escalate}: Escalation is triggered when the same category of action 
                has been attempted 3 or more consecutive times with no meaningful state change or 
                improvement in outcome. This threshold applies regardless of domain, whether it is 
                repeated failed movements, unsuccessful search queries, recurring test failures, or 
                fruitless object searches. At this point, persisting with the same approach is 
                counterproductive. The agent must halt, shift into diagnostic mode, and treat all 
                prior feedback from the environment as structured evidence rather than mere failure.
                
                \textbf{3. How to Break the Loop}: Upon detecting stagnation, the agent should first 
                articulate, within its reasoning step, exactly what was tried, what the environment 
                returned, and what gap in knowledge or strategy is causing the block. The agent should 
                then consider whether the problem has been correctly framed, whether an environmental 
                rule has been misunderstood, and whether any untried alternative exists. If external 
                tools or a knowledge base are available (like \texttt{search\_and\_ask}), they should be consulted immediately with a 
                precise, context-rich query rather than deferred. Action resumes only under a 
                substantively different strategy.
                
                \textbf{4. What to Avoid}: The agent must not treat repeated failure as a reason to 
                keep trying the same action, nor mistake tool availability for optional assistance, 
                tools exist to be used when direct attempts stall. Overconfidence in a single strategy, 
                blindness to deadlock states, and circular revisiting of already-exhausted options are 
                the primary failure modes this protocol is designed to prevent. Equally, the agent 
                should not escalate prematurely, slow progress is not stagnation, and a single failure 
                does not warrant a full strategy shift.
                
                \textbf{5. Underlying Principle}: Every failed attempt carries diagnostic value — it 
                narrows the solution space and reveals constraints. An intelligent agent does not 
                persist blindly but adapts purposefully: recognizing when a strategy has run its course, 
                identifying whether the obstacle stems from a knowledge gap or a wrong approach, and 
                pivoting with intent. This capacity to self-monitor, diagnose, and escalate is what 
                separates robust task completion from brittle, loop-prone behavior across any environment.
            \end{minipage}
        \end{tabular}
    \end{tcolorbox}
    \label{tab:multitask_exp}
    \end{minipage}
\end{table}

\end{document}